\documentclass[final]{cvpr}

\usepackage{times}
\usepackage{epsfig}
\usepackage{graphicx}
\usepackage{amsmath}
\usepackage{amssymb}

\usepackage{bm}
\usepackage{multirow}
\usepackage{dsfont}

\usepackage{xcolor}
\usepackage{wrapfig}
\usepackage{booktabs}
\usepackage{color}
\usepackage{verbatim}
\usepackage{caption}
\usepackage{ulem}
\usepackage{enumitem}
\usepackage{placeins}
\usepackage{ragged2e}
\usepackage{diagbox}
\usepackage{enumitem}

\usepackage{makecell}
\usepackage{subfigure}
\usepackage{arydshln}
\usepackage{diagbox}

\newcommand{\qy}[1]{\textcolor{black}{#1}}
\newcommand{\bo}[1]{\textcolor{black}{#1}}
\newcommand{\unclear}[1]{\textcolor{black}{#1}}
\newcommand{\nickname}{SpinNet}
\newcommand{\qysupp}[1]{\textcolor{black}{#1}}

\usepackage[pagebackref=true,breaklinks=true,colorlinks,bookmarks=false]{hyperref}

\def\cvprPaperID{2104} % *** Enter the CVPR Paper ID here
\def\confYear{CVPR 2021}

\begin{document}

%%%%%%%%% TITLE
\title{\nickname{}: Learning a General Surface Descriptor for 3D Point Cloud Registration}

\author{Sheng Ao\textsuperscript{1*}, Qingyong Hu\textsuperscript{2\thanks{Equal contribution}*}, Bo Yang\textsuperscript{3}, Andrew Markham\textsuperscript{2}, 
Yulan Guo\textsuperscript{1,4} \\
\textsuperscript{1}Sun  Yat-sen University, \textsuperscript{2}University of Oxford, \\
\textsuperscript{3}The Hong Kong Polytechnic University, \textsuperscript{4}National University of Defense Technology\\
{\tt\small qingyong.hu@cs.ox.ac.uk, bo.yang@polyu.edu.hk, guoyulan@mail.sysu.edu.cn }}

\maketitle

\begin{abstract}

\qy{Extracting robust and general 3D local features is key to downstream tasks such as point cloud registration and reconstruction. Existing learning-based local descriptors are either sensitive to rotation transformations, or rely on classical handcrafted features which are neither general nor representative. In this paper, we introduce a new, yet conceptually simple, neural architecture, termed \nickname{}, to extract local features which are rotationally invariant whilst sufficiently informative to enable accurate registration. A Spatial Point Transformer is first introduced to map the input local surface into a carefully designed cylindrical space, enabling end-to-end optimization with SO(2) equivariant representation. A Neural Feature Extractor which leverages the powerful point-based and 3D cylindrical convolutional neural layers is then utilized to derive a compact and representative descriptor for matching. Extensive experiments on both indoor and outdoor datasets demonstrate that \nickname{} outperforms existing state-of-the-art techniques by a large margin. More critically, it has the best generalization ability across unseen scenarios with different sensor modalities. The code is available at \url{https://github.com/QingyongHu/SpinNet}.}
\end{abstract}

%%%%%%%%% BODY TEXT

\section{Introduction}
\label{sec:Intro}
\begin{figure}[t]
	\begin{center}
		\includegraphics[width=1.0\linewidth]{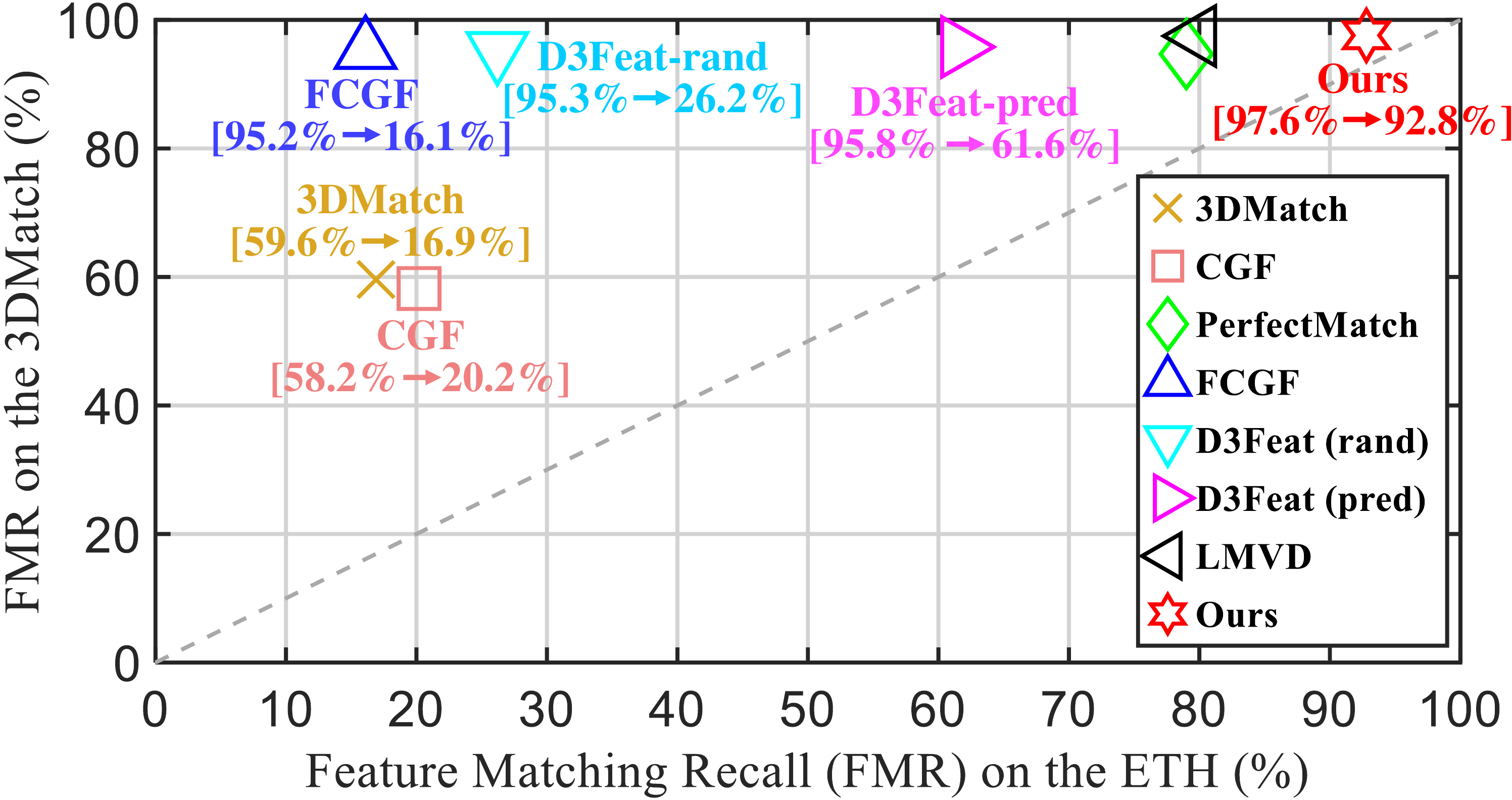}
	\end{center}
    \caption{\qy{The Feature Matching Recall (FMR) scores of different approaches on the indoor 3DMatch~\cite{Zeng2017} and outdoor ETH~\cite{pomerleau2012challenging} dataset. Note that, all methods are trained only on the 3DMatch dataset. Our method not only achieves the highest score on 3DMatch, but also has the best generalization ability across the unseen ETH dataset. 
    }}
	\label{fig: generalization}
\end{figure}

\qy{
\bo{Accurate matching of partial 3D surfaces is critical for} point cloud registration~\cite{Elbaz2017,Lu2019,Choy2020,Gojcic2020a,huang2020feature}, segmentation~\cite{yang2019learning,hu2020randla,guo2020deep}, and recognition~\cite{guo20143d,Defferrard2020,Rao2019}. Given multiple partially overlapped 3D scans, the goal of surface matching is to align these fragments according to a set of point correspondences, thus obtaining a complete 3D scene structure. \bo{To achieve this, it is of key importance to identify general and robust local geometric patterns shared between two scans. However, this is challenging, primarily because 1) different scans usually have different viewing angles, 2) the raw 3D scans are typically incomplete, noisy, and have significantly different point densities.} 
}

\qy{Early methods to extract local geometries include PS~\cite{chua1997point}, ISS~\cite{Zhong2010Intrinsic}, SHOT~\cite{mian2010repeatability} and RoPS~\cite{guo2013rotational}, which simply compute the low-level features such as faces~\cite{mian2006three,zaharescu2009surface}, corners~\cite{stein1992structural}, and handcrafted statistical histograms~\cite{novatnack2008scale}. \bo{Although they achieve encouraging results on high-quality 3D point clouds, they are not capable of generalizing to highly noisy and large-scale real-world 3D point clouds.}
}

\qy{
\bo{Recent deep neural network based approaches \cite{Zeng2017,Khoury2017,Deng2018,Poiesi2021} have yielded excellent results in learning better local point features, thanks to the availability of large-scale labeled 3D datasets. However, they have two major limitations. \textbf{First}, many of these methods such as D3Feat \cite{bai2020d3feat} and FCGF \cite{choy2019fully} rely on kernel-based point convolution \cite{kpconv} or submanifold sparse convolution \cite{sparseconv} to extract per-point features, resulting in the learned point local patterns being rotationally variant. Consequently, their performance drops dramatically when they are applied to novel 3D scans with strong rotational changes. \textbf{Second}, although a number of recent approaches \cite{Deng2018,Deng2018a,Gojcic2019,li2020end} introduce rotation-invariant point descriptors, they simply integrate the handcrafted features such as point-pairs \cite{rusu2009fast,drost2010model} and point density ~\cite{Spezialetti2019a,you2020pointwise,Spezialetti2020a}, or external local reference frames (LRFs) ~\cite{li2020end,Kim2020a,Zhao2020} into the pipeline, fundamentally limiting the representational power of the framework \cite{guo2016comprehensive}. As a result, the extracted point features, albeit being rotation invariant, are not robust and general when being applied to unseen 3D scans with noise and different point densities.}} %\cite{Zhao2020}

\bo{In this paper, we aim to design a new neural architecture, which is able to learn descriptive local features and generalize well to unseen scenarios. This network clearly satisfies three key properties: 1) It is rotation invariant. Particularly, it learns consistent local features from 3D scans with different rotation angles; 2) It is descriptive. In essence, it preserves the prominent local patterns despite the noise, possible surface incompleteness, or different point densities; 3) It does not include any handcrafted features. Instead, it only consists of multiple point transformations and simple neural layers coupled with true end-to-end optimization. This allows the learned descriptor to be extremely representative and general for complex real-world 3D surfaces.}

\bo{Our network, named \nickname{}, mainly consists of two modules, 1) a \textbf{Spatial Point Transformer}\footnote{This is different from the Transformer for natural language processing.}, which explicitly transforms the input 3D scans into a carefully designed cylindrical space, driving the transformed scans to be SO(2) equivariant, whilst retaining point local information; 2) a \textbf{Neural Feature Extractor}, which leverages powerful point-based and convolutional neural layers to learn representative and general local patterns.}

\bo{The \textbf{Spatial Point Transformer} firstly aligns the input 3D surface according to a reference axis, eliminating the rotational variance along the Z-axis. This is followed by an additional coordinate transformation over the XY-plane with the aid of spherical voxelization, further removing the rotation variance of each spherical voxel. Lastly, the transformed local surface is formulated as a simple yet novel 3D cylindrical volume, which is amenable to consumption by the subsequent point-based and convolutional neural layers. The \textbf{Neural Feature Extractor} firstly uses simple point-based MLPs to extract a unique signature for each voxel within the cylindrical volume, generating an initial set of cylindrical feature maps. These maps are further fed into a series of novel 3D cylindrical convolutional layers, which fully exploit the rich spatial and contextual information and generate a compact and representative feature vector.}
% for the input 3D surface

\qy{
\bo{These two modules enable our \nickname{} to learn remarkably robust and general local features for accurate 3D point cloud registration.}
It achieves state-of-the-art performance both on the indoor 3DMatch ~\cite{Zeng2017} dataset and the outdoor ETH \cite{pomerleau2012challenging} dataset. Notably, it shows superior generalization ability across unseen scenarios. As shown in Figure \ref{fig: generalization}, being trained only on the 3DMatch dataset, the learned descriptor of our \nickname{} can achieve an average recall score of 92.8\% on the unseen outdoor ETH dataset for feature matching, significantly surpassing the state of the art by nearly 13\%. Overall, our contributions are three-fold:
}
\begin{itemize}[leftmargin=*]
\setlength{\itemsep}{0pt}
\setlength{\parsep}{0pt}
\setlength{\parskip}{0pt}
\item \qy{We propose a new neural feature learner for 3D surface matching. It is rotation invariant, representative, and has superior generalization ability across unseen scenarios.}
\item \qy{By formulating the transformed 3D surface into a cylindrical volume,  
we introduce a powerful 3D cylindrical convolution to learn rich and general features.}
\item \qy{We conduct extensive experiments and ablation studies, demonstrating the remarkable generalization of our method and providing the intuition behind our choices.}
\end{itemize}

\section{Related Work}
\label{sec:related_work}

\subsection{Handcrafted Descriptors}
\qy{Traditional handcrafted descriptors can be roughly divided into two categories: 1) LRF-free methods and 2) LRF-based. The LRF-free descriptors including Spin Images (SIs)~\cite{johnson1999using}, Local Surface Patch (LSP)~\cite{chen20073d} and Fast Point Feature Histograms (FPFHs)~\cite{rusu2009fast} are typically constructed by exploiting geometrical properties (\textit{e.g.} curvatures and normal deviations) of a local surface. The main drawback of these descriptors is the lack of sufficient geometric details for the local surface. The LRF-based descriptors such as Point Signature (PS)~\cite{chua1997point}, SHOT~\cite{tombari2010unique} and Rotational Projection Statistics (RoPS)~\cite{guo2013rotational} are not only able to characterize the geometric patterns of the local support region, but also effectively exploit the 3D spatial attributes. However, LRF-based methods inherently introduce rotation errors, sacrificing the feature robustness. Overall, all these handcrafted descriptors are usually tailored to specific tasks and sensitive to noise, thus not being sufficiently flexible and descriptive for complicated and novel scenarios. }

\subsection{Learning-based Descriptors}
\qy{In contrast to traditional handcrafted descriptors, recent works~\cite{Cohen2018,Esteves2018,Zhou2018, Chen2019,Yu2020,yew2020rpm} leverage data-driven deep neural networks to learn local features from large-scale datasets. These learned descriptors tend to have strong descriptive ability and robustness.} 

\textbf{Rotation Variant Descriptors.} \qy{ Zeng \textit{et al.} propose the pioneering work 3DMatch ~\cite{Zeng2017}, which takes the local volumetric patches as input, and then leverages 3D Convolutional Neural Networks (CNNs) to learn local geometric patterns. Yew and Lee introduce a weakly-supervised framework 3DFeat-Net ~\cite{Yew2018} to learn both the 3D feature detector and descriptor simultaneously. Choy \textit{et al.} build a dense feature descriptor FCGF ~\cite{choy2019fully} based on~\cite{choy20194d}. Recently, Bai \textit{et al.}~\cite{bai2020d3feat} design a pipeline to jointly learn both dense feature detectors and local feature descriptors, achieving the state-of-the-art performance on 3DMatch \cite{Zeng2017} and KITTI~\cite{geiger2012we} datasets for point cloud registration.}
\qy{However, all these methods are sensitive to rigid transformation in Euclidian space. Extensive data augmentation  can be used to alleviate this problem, however, the overall performance of subsequent tasks is still sub-optimal ~\cite{Esteves2018}.}
%(\textit{e.g.}, arbitrary rotation)

\textbf{Rotation Invariant Descriptors.} 
\qy{A number of recent methods have started to learn rotation-invariant descriptors. Khoury \textit{et al.}~\cite{Khoury2017} parameterize the raw point clouds with oriented spherical histograms, and then map the high-dimensional embedding to a compact descriptor through a deep neural network. Deng \textit{et al.}~\cite{Deng2018} encode the local surface using rotation-invariant Point Pair Features (PPFs). These features are then fed into multiple MLPs to learn a global descriptor. In the follow-up work~\cite{Deng2018a}, FoldingNet~\cite{Yang2018} is adopted as the backbone network to learn 3D local descriptors. Gojcic \textit{et al.}~\cite{Gojcic2019} introduce the voxelized Smoothed Density Value (SDV) to encode the local surface as a compact and rotation-invariant representation, which is fed into a Siamese architecture to learn the final descriptor.}
\qy{Overall, although these methods are indeed able to learn rotationally invariant features from the local surface, they initially rely on classical handcrafted features which significantly limits the descriptiveness of the descriptors. Additionally, most of the above handcrafted features are based on the point-pairs and point density, both of which are sensitive to noise, clutter, and distance variations, making the learned features hardly generalize to novel scenarios.}

\qy{A handful of recent works~\cite{Spezialetti2019a,you2020pointwise,Kim2020a,li2020end} try to learn rotation-invariant local descriptors with end-to-end optimization. However, they either require the computation of the point density or rely on external reference frames to achieve rotation invariance. This is usually unstable and does not generalize well to unseen datasets. 
In contrast, our \nickname{} directly transforms the point clouds into a cylindrical volume followed by a series of powerful neural layers. It learns rotation-invariant, compact, and highly descriptive local features in a truly end-to-end fashion, without relying on any handcrafted features or unstable external LRFs. This enables the learned descriptors to be extremely general for use on unseen 3D surfaces across different datasets.}

\section{\nickname{}}

\begin{figure*}
	\begin{center}
		\includegraphics[width=1.0\linewidth]
		{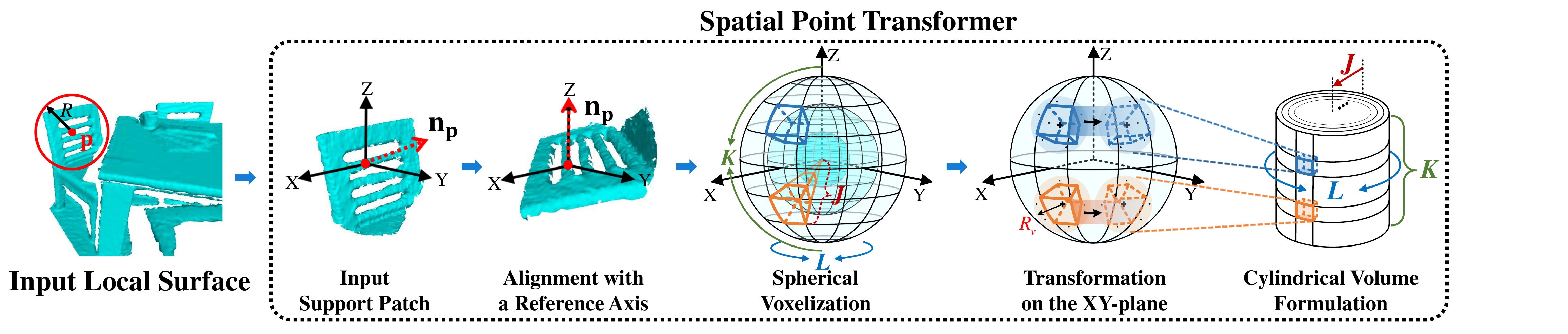}
	\end{center}
	\caption{\qy{The detailed components and processing steps of our Spatial Point Transformer.}}
	\label{fig: parameterization}
\end{figure*}

\subsection{Problem Statement}
\qy{Given two partially overlapped point clouds  $\mathcal{P} = \{p_i \in \mathbb{R}^{3}| i= 1,\dots, N \}$ and $\mathcal{Q} = \{q_j \in \mathbb{R}^{3}| j= 1,\dots, M \}$. The task of point cloud registration is to find an optimal rigid transformation $\mathbf{T} = \{ \mathbf{R}, \mathbf{t} \}$, as well as the point correspondences to align pairs of fragments, and finally recover the complete scene. The pair of point correspondence $({p}_i, {q}_j )$ is expected to satisfy:
}
\begin{equation} \label{eq1}
{q}_j = \mathbf{R}{p}_i + \mathbf{t} + \mathbf{\epsilon}_i,
\end{equation}
\qy{where $\mathbf{R} \in \text{SO(3)}$ denotes the rotation matrix,  $\mathbf{t} \in \mathbb{R}^{3}$ is the translation vector, and $\mathbf{\epsilon}_i$ is the residual error. In practice, it is infeasible to simultaneously find the correspondences and estimate the transformation, due to the non-convexity of this problem~\cite{li20073d}. However, if the point subsets $\mathcal{P} ^c$ and $\mathcal{Q} ^c$ with one-to-one correspondences can be determined, the registration problem can be simplified as a  minimization problem for the following $L_{2}$ distance:
} 
\begin{equation} \label{eq2}
\mathcal{L}(\mathcal{P}^ c, \mathcal{Q}^ c|\mathfrak{P}, \mathbf{R},  \mathbf{t} ) = \frac{1}{\mathcal{N}} \left \| \mathcal{Q}^ c - \mathbf{R}  \mathcal{P}^ c \mathfrak{P}  -  \mathbf{t} \right \|^{2}
\end{equation}
\qy{where $\mathcal{N}$ is the number of successfully matched correspondences, $\mathfrak{P}\in \mathbb{R}^{\mathcal{N} \times \mathcal{N}}$ 
is a permutation matrix whose entries satisfy $\mathfrak{P}_{u,v} = 1$ if the $u^{th}$ point in $\mathcal{P} ^c$ corresponds to $v^{th}$ point in $\mathcal{Q} ^c$ and 0 otherwise.}

\qy{We propose a new surface feature learner \nickname{}, which is a mapping function $\mathcal{M}$, where $\mathcal{M}({p}_i) $ is equal to $\mathcal{M}({q}_j)$ under arbitrary rigid transformations such as rotation and translation, if ${p}_i$ and ${q}_j$ are indeed a correct match. In particular, our feature learner mainly consists of a Spatial Point Transformer and a Neural Feature Extractor.}

\subsection{Spatial Point Transformer}
\label{subsec: FEL}

\bo{This module is designed to spatially transform the input 3D surfaces into a cylindrical volume, overcoming the rotation variance, whilst without dropping critical information of local patterns. As shown in Figure ~\ref{fig: parameterization}, it consists of four components, as discussed below.}

\textbf{Alignment with a Reference Axis.} \qy{Given a specific point $\mathbf{p}\in \mathcal{P}$ in a local surface, we first estimate a reference axis $\mathbf{n}_\mathbf{p}$ oriented to the viewpoint~\cite{mian2010repeatability,ao2020repeatable} from its neighbouring point set $\mathbf{P}^\mathrm{s} = \{ \mathbf{p}_i: \left\|\mathbf{p}_i-\mathbf{p}\right\|^2 \leq R\}$ within a support radius $R$. We then align $\mathbf{n}_\mathbf{p}$ with the Z-axis using a rotation matrix $\mathbf{R}_\mathbf{z}$. Compared with the external local reference frames which are likely to be ambiguous and unstable, our estimated $\mathbf{n}_\mathbf{p}$ tends to be more robust and stable with regard to rotation changes \cite{petrelli2011repeatability}.}
\qy{Subsequently, the neighbouring point set $\mathbf{P}^\mathrm{s}$ is transformed to $\mathbf{P}^\mathrm{s}_\mathrm{r} = \mathbf{R}_\mathbf{z}\mathbf{P}^\mathrm{s}$.}
\qy{
To achieve translation invariance, we further normalize $\mathbf{P}^\mathrm{s}_\mathrm{r}$ by offsetting to the center point, \textit{i.e.}, $\mathbf{\widehat{P}}^\mathrm{s}_\mathrm{r} = \mathbf{P}^\mathrm{s}_\mathrm{r} - \mathbf{R}_\mathbf{z}\mathbf{p}$. Hence, the obtained local patch $\mathbf{\widehat{P}}^\mathrm{s}_\mathrm{r}$ is aligned with the $\mathbf{z}$-axis, leaving the remaining rotational degree of freedom entirely on the XY-plane.}

\textbf{Spherical Voxelization.} \qy{
To further eliminate the rotational variance on the XY-plane, we leverage a rotation-robust spherical representation. In particular, we treat the patch $\mathbf{\widehat{P}}^\mathrm{s}_\mathrm{r}$ as a sphere, and evenly divide it into $ J \times K \times L$ voxels along the radial distance $\rho$, elevation angle $\phi$ and azimuth angle $\theta$. The center of each voxel is denoted as $\mathbf{v}_{jkl}$, where $j \in \{1,...,J\}$, $k \in\ \{1,...,K\}$, $l \in \{1,...,L\}$. We then explicitly identify a set of neighboring points for the center point $\mathbf{v}_{jkl}$ of each voxel. In particular, we use the radius query to find the neighboring points $\mathbf{P}_{jkl} \subset \mathbf{\widehat{P}}^\mathrm{s}_\mathrm{r}$ based on a fixed radius $R_v$, where $\mathbf{P}_{jkl} = \{ \mathbf{\widehat{p}}_i :\left\|\mathbf{\widehat{p}}_i-\mathbf{v}_{jkl}\right\|^2 \leq R_v, \mathbf{\widehat{p}}_i \in \mathbf{\widehat{P}}^\mathrm{s}_\mathrm{r} \}
$. Lastly, we randomly sample and preserve a fixed number of $k_v$ points for each voxel, aiming for efficient computation in parallel. This spherical voxelization step is key to the successive spatial point transformation.}

\textbf{Transformation on the XY-Plane.} \qy{To enable each spherical voxel to be rotationally invariant on the XY-plane,
we proactively rotate each voxel around the Z-axis to align its center $\mathbf{v}_{jkl}$ with the YZ-plane, where the rotation matrix $\mathbf{R}_{jkl}$ is defined as:}
\begin{equation} \label{eq3} 
\mathbf{R}_{jkl} = \begin{bmatrix}
\cos(\pi/2-2\pi l/L) & -\sin(\pi/2 - 2\pi l/L) & 0 \\
\sin(\pi/2-2\pi l/L) & \cos(\pi/2-2\pi l/L) & 0 \\
0 & 0 & 1
\end{bmatrix}
\end{equation}
\qy{This removes an additional rotational degree of freedom for each voxel on the XY-plane, without dropping any local geometric patterns of each voxel. Note that, the existing methods \cite{Spezialetti2019a,you2020pointwise} usually use handcrafted features to achieve rotation invariance, resulting in the loss of the rich local patterns. Uniquely, our simple strategy to transform voxels can preserve these patterns, leaving them to be learned by the powerful neural layers.
}

{\bf Cylindrical Volume Formulation.} \qy{
\bo{Once the local patterns of each voxel are transformed, it is crucial to further preserve the larger spatial structures across multiple voxels. This requires the relative positions of all voxels to be represented in the whole framework. 
To this end, we reformulate the spherical voxels into a cylindrical volume. This is amenable to the proposed 3D cylindrical convolutional network, which guarantees the SO(2) equivariance  of the input local surface and preserves the topological patterns of multiple voxels.}}
\bo{In particular, given the transformed spherical voxels, each of which has a set of neighbouring points, we logically project them into a cylindrical volume, denoted as $\mathbf{C} \in \mathbb{R}^{J \times K \times L \times k_v \times 3}$ and illustrated in Figure \ref{fig: parameterization}.
}

\bo{In summary, given an input surface patch, our Spatial Point Transformer explicitly aligns its Z-axis with a reference axis, and proactively transforms the spherical voxel patterns on the XY-plane, and further preserves the topological surface structures through the cylindrical volume formulation. Clearly, this module keeps all surface patterns intact for the subsequent Neural Feature Extractor to learn.}

\begin{figure*}[thb]
	\begin{center}
		\includegraphics[width=1.0\linewidth]{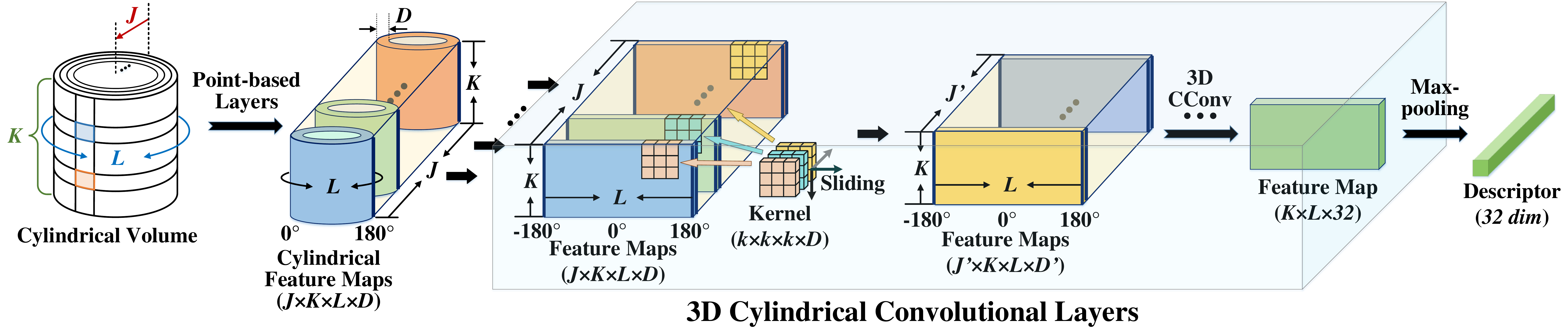}
	\end{center}
	\caption{\qy{Illustration of the proposed Neural Feature Extractor.}}
	\label{fig: 3DCCN}
\end{figure*}
\vspace{-0.2cm}

\subsection{Neural Feature Extractor}
\label{subsec: CCN}
\bo{This module is designed to learn the general features from the transformed points within each cylindrical voxel using the powerful neural layers. As shown in Figure \ref{fig: 3DCCN}, it consists of two components, as discussed below.}

{\bf Point-based Layers.} \qy{ Given the points within each cylindrical voxel, we use shared MLPs followed by a max-pooling function $\mathcal{A}(\cdot)$ to learn an initial signature for each voxel. Formally, the point-based layers are defined as:}
\begin{equation} \label{eq4} 
\mathbf{f}_{jkl} =\mathcal{A}(\mathbf{MLPs}(\mathbf{R}_{jkl}\mathbf{P}_{jkl}))
\end{equation}
\qy{where $\mathbf{f}_{jkl}$ is the learned features with $D$ dimension. Note that, the MLP weights are shared across all spherical voxels. Eventually, we obtain a set of 3D cylindrical feature maps $\mathbf{F} \in \mathbb{R}^{J \times K \times L \times D}$. 
}

\bo{\bf 3D Cylindrical Convolutional Layers.}
\qy{To further learn spatial structures across multiple voxels of the volume, we propose an efficient 3D Cylindrical Convolution Network (3DCCN) inspired by \cite{joung2020cylindrical}. In particular, given a voxel located at the position $(j,k,l)$ on the $d^{th}$ cylindrical feature map in the $s^{th}$ layer, our 3DCCN is defined as follows. }
\begin{equation} \label{eq10}
\begin{split}
\mathbf{F}_{jkl}^{sd'}=\sum_{d=1}^{D}\sum_{r=1}^{R_s}\sum_{y=1}^{Y_s}\sum_{x=1}^{X_s} w_{ryx}^{sd'd} \mathbf{F}_{(j+r)(k+y)(l+x)}^{(s-1)d}.
\end{split}
\end{equation}
where $R_s$ is the size of the kernel along the radial dimension, $Y_s$ and $X_s$ are the height and width of the kernel respectively, $w_{ryx}^{sd'd}$ are the learnable parameters. 

\bo{Being quite different from existing convolution operations, our proposed 3DCCN is novel in the following two aspects. \textbf{First}, since the cylindrical feature maps are $360^{\circ}$ continuous over a cylinder, our 3DCCN is designed to wrap around these feature maps, \textit{i.e.,} over the periodic boundary from $-180^{\circ}$ to $180^{\circ}$. Therefore, explicit padding is not required in our 3DCCN, but required by 3D-CNN at the boundary of feature maps. \textbf{Second}, compared with the existing 3D manifold sparse convolution \cite{choy2019fully} or kernel point convolutions \cite{bai2020d3feat}, the continuous convolution around the $360^{\circ}$ volume enables the obtained feature map to be SO(2) equivariant, hence to achieve the  final rotation-invariance.
}

After stacking multiple of these 3DCCN layers followed by max-pooling, the original cylindrical feature maps are compressed to a compact and representative feature vector.

\begin{comment}
\qy{As shown in Figure~\ref{fig: 3DCCN}, the 3D cylindrical space is defined as $S \times H \times \Theta$, we then formulate the 3D cylindrical feature maps and filters as continuous functions $f: S \times H \times \Theta \rightarrow \mathbb{R}^{D}$, where $S$, $H$ and $\Theta$ represent the radial distance to the center, height, and azimuth angle of cylindrical signal respectively. $D$ is the number of channels. Formally, 3D cylindrical convolution takes the cylinder signal $f$ as input and convolves it with the filter $\psi: S \times H \times \Theta \rightarrow \mathbb{R}^{D}$:}
\begin{equation} \label{eq9}
\begin{split}
&[f\ast\psi](s,h,\theta)= \\
&\int_{S}\int_{H}\int_\Theta{\sum_{d=1}^{D}f_{d}(x,y,z)\psi_{d}(x-s,y-h,z-\theta)}dzdydx.
\end{split}
\end{equation}
\end{comment}

\subsection{End-to-end Implementation}
\label{subsec: network}
\qy{ The Spatial Point Transformer is directly connected with the Neural Feature Extractor, followed by the existing contrastive loss \cite{bai2020d3feat} for end-to-end optimization. The widely-used \textit{hardest in batch} sampling~\cite{mishchuk2017working} is also adopted on-the-fly to maximize the distance between the closest positive and the closest negative patches. Details of the neural layers are presented in the appendix.}

\qy{We implement our \nickname{} based on the PyTorch framework. The Adam optimizer \cite{kingma2014adam} with default parameters is used. The initial learning rate is set to 0.001 and decayed with a rate of 0.5 for every 5 epochs. We train the network for 20 epochs, the best-performed model on the validation set is then used for testing. For a fair comparison, we keep the same setting for all experiments. All experiments are conducted on the platform with Intel Xeon CPU @2.30GHZ with an NVIDIA RTX2080Ti GPU.}

\begin{comment}
{\bf Architecture.} 
\qy{We stack several cylindrical convolutional layers together to learn a more descriptive feature representation with SO(2) rotation-equivariance. In addition, we follow R2D2~\cite{Revaud2019} to stack three 2$\times$2 cylindrical convolution layers together to reduce the parameters and computational cost. The last feature maps are further aggregated using a non-parametric max-pooling layer, and the final output is a rotation-invariant and compact (32-dim) local feature descriptor. To avoid overfitting, the whole framework is lightweight with a small number of trainable parameters (about 2 million), which is far fewer than FCGF \cite{choy2019fully} and D3Feat \cite{bai2020d3feat}. For the detailed network architecture, please refer to the supplementary materials.}
\end{comment}

\section{Experiments}
\label{Experiments}

We first evaluate our \nickname{} on the indoor 3DMatch dataset \cite{Zeng2017}  and the outdoor KITTI dataset \cite{geiger2012we}. We then evaluate the generalization ability of our approach across multiple unseen datasets \cite{Zeng2017, geiger2012we, pomerleau2012challenging} acquired by different sensors. Lastly, extensive ablation studies are conducted.

\textbf{Experimental Setup.} \qy{We follow~\cite{bai2020d3feat, Zeng2017} to generate training samples by only considering the point cloud fragment pairs with more than 30\% overlap in the whole dataset. For each paired fragment $\mathbf{P}$ and $\mathbf{Q}$, we randomly sample a fixed number of anchor points from the overlapping region of $\mathbf{P}$, and then apply the ground-truth transformation $\mathbf{T} = \{ \mathbf{R}, \mathbf{t} \}$ to determine the corresponding points in fragment $\mathbf{Q}$. Considering the varying number of point cloud fragments in different datasets, we uniformly select 20 and 500 anchor points from each fragment in the 3DMatch and KITTI dataset, so as to keep a similar number of samples for training. For each anchor point, we randomly sample 2048 points from its support region. }

\subsection{Evaluation on Indoor 3DMatch Dataset}
\label{Ex:3dmatch}

\qy{3DMatch is a RGBD-reconstruction dataset, which consists of 62 real-world indoor scenes collected from existing dataset~\cite{valentin2016learning, shotton2013scene, xiao2013sun3d, kim2014shape2pose, valentin2016learning, dai2017bundlefusion}. We follow the official protocol provided in~\cite{bai2020d3feat} to divide the scenes into training and testing splits. Each scene contains several partially overlapped fragments, and has the ground truth transformation parameters available for evaluation. Feature Matching Recall (FMR) \cite{Deng2018a} is used as the standard metric.
}

\smallskip\noindent\textbf{Comparisons with the state-of-the-arts.} \qy{ We first compare the FMR scores achieved by our \nickname{} and strong baselines (including LMVD \cite{li2020end}, D3Feat \cite{bai2020d3feat}, FCGF \cite{choy2019fully}, PerfectMatch \cite{Gojcic2019}, PPFNet \cite{Deng2018}, and PPF-FoldNet \cite{Deng2018a}) on the 3DMatch dataset, under the conditions of sampling points $f$=5000, distance threshold $\tau_1$=10 cm and inlier ratio threshold $\tau_2$=5\%. To further evaluate the robustness of all approaches against rotations, we follow \cite{Deng2018, bai2020d3feat} to build a rotated 3DMatch benchmark by applying arbitrary rotations in SO(3) group to all fragments of the dataset. 
}

\begin{table}[htb]
	\begin{center}
		\scalebox{0.75}{
			\begin{tabular}{r|cc|cc|c|c}
		       \Xhline{2\arrayrulewidth}
				\multirow{2}{*}{} & \multicolumn{2}{c|}{\bf{Origin}}  & \multicolumn{2}{c|}{\bf{Rotated}} & {Feat.} & {Rot.} \\
				& FMR (\%)      & STD         & FMR (\%)      & STD   & {dim.}   & {Aug.}    \\ \hline
				FPFH~\cite{rusu2009fast}         & 35.9          & 13.4         & 36.4          & 13.6  & {33}    & {No}    \\
				SHOT~\cite{tombari2010unique}         & 23.8          & 10.9         & 23.4          & 9.5  & {352}     & {No}     \\
				3DMatch~\cite{Zeng2017}      & 59.6          & 8.8          & 1.1           & -       & {512}     & {No}      \\
				CGF~\cite{Khoury2017}          & 58.2          & 14.2         & 58.5          & 14.0   & {32}   & {No}   \\
				PPFNet~\cite{Deng2018}       & 62.3          & 10.8         & 0.3           & -        & {64}      & {No}     \\
				PPF-FoldNet~\cite{Deng2018a}  & 71.8          & 10.5         & 73.1          & 10.4    & {512}    & {No}   \\
				PerfectMatch~\cite{Gojcic2019} & 94.7          & 2.7          & 94.9          & 2.5    & {32}  & {No}    \\
				FCGF~\cite{choy2019fully}         & 95.2          & 2.9          & 95.3          & 3.3    & {32}      & {Yes}    \\
				D3Feat-rand~\cite{bai2020d3feat} & 95.3          & 2.7          & 95.2          & 3.2   & {32}    & {Yes}     \\
				D3Feat-pred~\cite{bai2020d3feat} & 95.8          & 2.9          & 95.5          & 3.5   & {32}     & {Yes}     \\
				LMVD~\cite{li2020end} & 97.5          & 2.8          & 96.9          & -   & {32}   & {No}     \\
				\textbf{\nickname{} (Ours)}         & \textbf{97.6} & \textbf{1.9} & \textbf{97.5} & \textbf{1.9} & \textbf{32} & {\textbf{No}} \\ 
		        \Xhline{2\arrayrulewidth}
			\end{tabular}
		}
	\end{center}
	\caption{\qy{ Quantitative results on the 3DMatch dataset, STD: standard deviation. The symbol `-' means the results are unavailable or STD under low FMRs ($<$10\%).}}
	\label{tab:3dmatch}
\end{table}

\begin{table}[thb]
	\begin{center}
	\scalebox{0.8}{
	\begin{tabular}{rcccccc}
        \Xhline{2\arrayrulewidth}
		\#\textbf{Sampled points}      & \textbf{5000}     & \textbf{2500}     & \textbf{1000} & \textbf{500} & \textbf{250}   & \textbf{Average} \\ \hline
		\multicolumn{7}{c}{Feature Matching Recall (\%)}         \\ \hline
		PerfectMatch~\cite{Gojcic2019}     & 94.7     & 94.2     & 92.6   & 90.1  & 82.9   & 90.9   \\
		FCGF~\cite{choy2019fully}       & 95.2     & 95.5     & 94.6  & 93.0  & 89.9  & 93.6       \\
		D3Feat-rand~\cite{bai2020d3feat}     & 95.3     & 95.1     & 94.2  & 93.6  & 90.8  & 93.8       \\
		D3Feat-pred~\cite{bai2020d3feat}     & 95.8     & 95.6     & 94.6 & 94.3 & 93.3  & 94.7       \\
		\textbf{\nickname{} (Ours)}             & \textbf{97.6}      &    \textbf{97.5}      & \textbf{97.3}  & \textbf{96.3} & \textbf{94.3}      & \textbf{96.6}       \\ 
        \Xhline{2\arrayrulewidth}
	\end{tabular}
	}
	\end{center}
\caption{\qy{Quantitative results on the 3DMatch dataset using different numbers of sampled points.}}
\label{tab:3dmatch_keypoints}
\end{table}

\begin{figure}[thb]
	\begin{center}
		\includegraphics[width=1.0\linewidth]{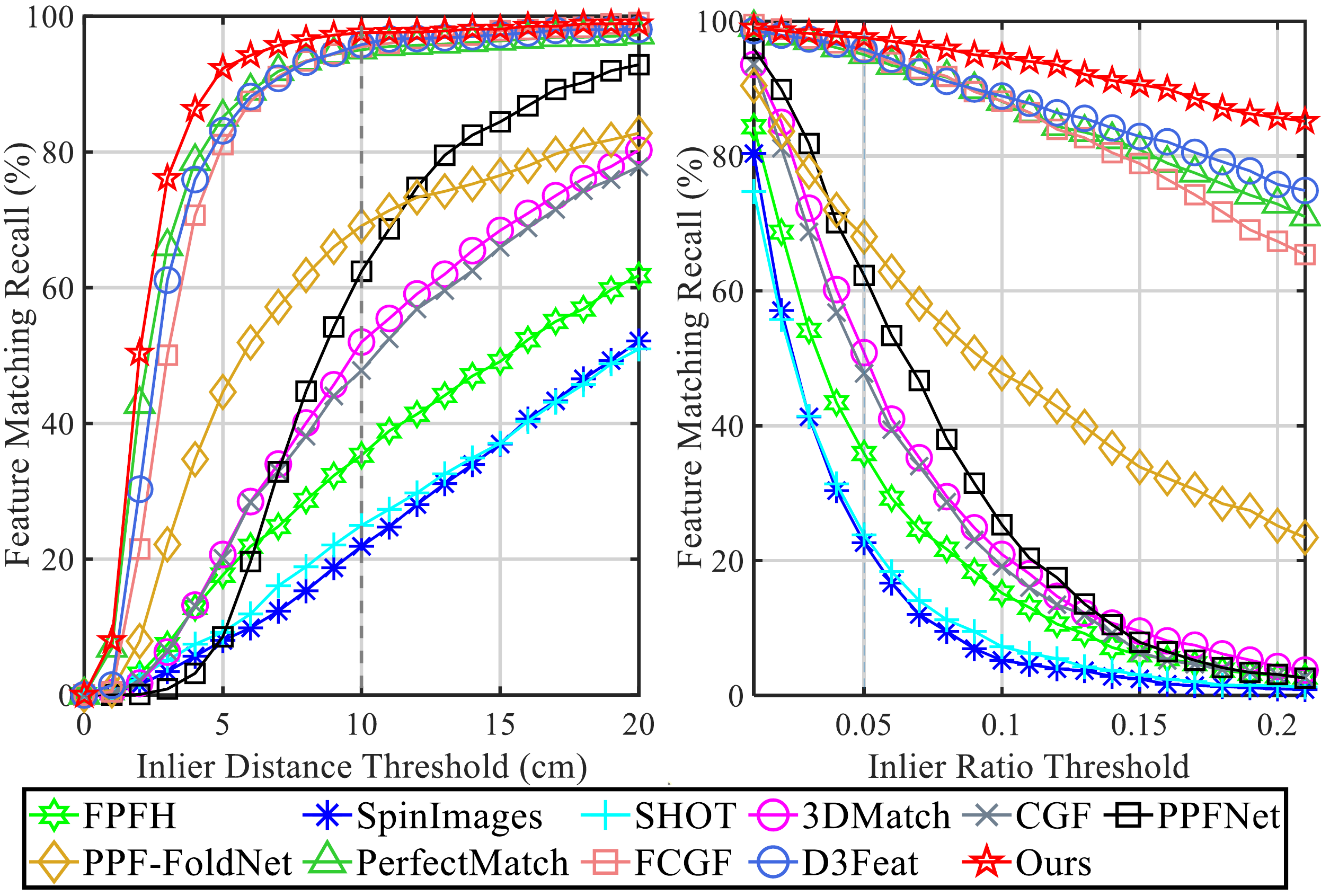}
	\end{center}
	\caption{\qy{Feature matching recall on the 3DMatch dataset under different inlier distance threshold $\tau_1$ (Left) and inlier ratio threshold $\tau_2$ (Right).}}
	\label{fig:3dmatch_FMR}
\end{figure}

\qy{As shown in Table~\ref{tab:3dmatch}, the descriptor generated by our method achieves the highest average FMR score and the lowest standard deviation on both the original and rotated datasets, outperforming the state-of-the-art methods. Note that, several baselines \cite{choy2019fully, bai2020d3feat} require rotation-based data augmentation for training, whilst ours does not.}

\smallskip\noindent\textbf{Performance under different number of sampled points.} \qy{ We further evaluate the performance of our \nickname{} on the 3DMatch by taking different number of sampled points as input. As shown in Table \ref{tab:3dmatch_keypoints}, the descriptor learned by our \nickname{} consistently achieves the best FMR scores when the number of sampled points is reduced from 5000 to 250. In particular, by randomly selecting points, our method even outperforms D3Feat-pred which has an explicit keypoint detection module. This demonstrates our network is highly robust and not sensitive to the number of sampled points.
}

\begin{table}[thb]
	\begin{center}
	\scalebox{0.8}{
	\begin{tabular}{r|cc|cc|c}
        \Xhline{2\arrayrulewidth}
		\multirow{2}{*}{} & \multicolumn{2}{c|}{\textbf{RTE (cm)}} & \multicolumn{2}{c|}{\textbf{RRE ($^\circ$)}} & \multirow{2}{*}{\textbf{Success (\%)}} \\
		& AVG                & STD               & AVG               & STD              &                       \\ \hline
		3DFeat-Net~\cite{Yew2018} & 25.9               & 26.2              & 0.57              & 0.46             & 95.97                 \\
		FCGF~\cite{choy2019fully}       & 9.52               & 1.30              & 0.30              & 0.28             & 96.57                 \\
		D3Feat-rand~\cite{bai2020d3feat}       & 8.78                   & 0.44                  & 0.32                  & 0.07                 & 99.81                      \\
		D3Feat-pred~\cite{bai2020d3feat}     & \textbf{6.90}               & \textbf{0.30}              & \textbf{0.24}              & \textbf{0.06}             & \textbf{99.81}                 \\
		\textbf{\nickname{} (Ours)}      & 9.88               & 0.50                  & 0.47                  & 0.09                 & 99.10                      \\ 
        \Xhline{2\arrayrulewidth}
	\end{tabular}
	}
	\end{center}
\caption{\qy{Quantitative results of different approaches on the KITTI odometry dataset. The scores of baselines are retrieved from ~\cite{bai2020d3feat}.}}
  \label{tab:kitti}
\end{table}

\smallskip\noindent\textbf{Performance under Different Error Thresholds.} \qy{ Additionally, we evaluate the robustness of \nickname{} by varying the error thresholds ($\tau_1$ and $\tau_2$). As shown in Figure ~\ref{fig:3dmatch_FMR}, the descriptor generated by \nickname{} consistently outperforms other methods under all thresholds. It is worth noting that the FMR score of our method is significantly higher than others, when the inlier ratio threshold increases. For a stricter condition $\tau_2$ = 0.2, our method maintains a high FMR score of 85.7\%, while D3Feat and FCGF drop to 75.8\% and 67.4\%, respectively. This highlights that our approach is more robust in harder scenarios.}

\subsection{Evaluation on Outdoor KITTI Dataset}
\label{Ex:kitti}

\qy{KITTI odometry \cite{geiger2012we} is an outdoor sparse point cloud dataset acquired by Velodyne-64 3D LiDAR scanners. It consists of 11 sequences of outdoor scans. For fair comparison, we follow the same dataset splits and preprocessing methods as used in D3Feat~\cite{bai2020d3feat, choy2019fully}. Similar to ~\cite{ma2016fast}, Relative Translational Error (RTE), Relative Rotation Error (RRE), and Success rate are used as the evaluation metrics. The registration is regarded as successful if the RTE and RRE of a pair of fragments are both below the predefined thresholds 2m and 5$^\circ$. It is noted that the point clouds are gravity-aligned in this dataset, we follow \cite{Yew2018} to skip the alignment with a reference axis in our method. As shown in Table \ref{tab:kitti}, the results of our \nickname{} are on par with the strong baseline D3Feat. Admittedly, our \nickname{} is marginally lower than the state-of-the-art D3Feat-pred, primarily because D3Feat has a powerful joint learned descriptor and keypoint detector. Also, the well aligned point clouds in this dataset are indeed in favor of D3Feat. We leave the integration of keypoint detection for future exploration.}

\subsection{Generalization across Unseen Datasets}
\label{Ex:generalization}

\qy{We have conducted several groups of experiments to extensively evaluate the generalization ability of our \nickname{}. In each group, our network is trained on one dataset, and then directly tested on a completely unseen dataset.}

\begin{table}[thb]
	\begin{center}
		\scalebox{0.7}{
			\begin{tabular}{r|c|cc|cc|c}
                \Xhline{2\arrayrulewidth}
				\multirow{2}{*}{} &
				{\bf{Param.}} &
				\multicolumn{2}{c|}{\bf{Gazebo}}   & \multicolumn{2}{c|}{\bf{Wood}}     & \multirow{2}{*}{Avg.}\\
				&(Mb)  & Summer        & Winter        & Autumn        & Summer        &                          \\ \hline
				FPFH$^\dagger$~\cite{rusu2009fast}                & -   & 38.6          & 14.2          & 14.8          & 20.8          & 22.1                     \\
				SHOT$^\dagger$~\cite{tombari2010unique}          & -        & 73.9          & 45.7          & 60.9          & 64.0          & 61.1                     \\
				3DMatch~\cite{Zeng2017} & 13.40     & 22.8          & 8.3           & 13.9          & 22.4          & 16.9                     \\
				CGF~\cite{Khoury2017} & 1.86     & 37.5          & 13.8          & 10.4          & 19.2          & 20.2                     \\
				PerfectMatch~\cite{Gojcic2019} & 3.26  & \underline{91.3}          & \underline{84.1}          & 67.8          & 72.8          & 79.0                     \\
				FCGF~\cite{choy2019fully} & 33.48  & 22.8          & 10.0          & 14.8          & 16.8          & 16.1                     \\
				D3Feat (rand)~\cite{bai2020d3feat} & 13.42  & 45.7          & 23.9          & 13.0          & 22.4          & 26.2                     \\
				D3Feat (pred)~\cite{bai2020d3feat} & 13.42  & 85.9          & 63.0          & 49.6          & 48.0          & 61.6                     \\
				LMVD~\cite{li2020end} & 2.66  & 85.3          & 72.0          & \underline{84.0}          & \underline{78.3}          & \underline{79.9}                     \\			
				\textbf{\nickname{} (Ours)} & 2.16  & \textbf{92.9} & \textbf{91.7} & \textbf{92.2} & \textbf{94.4} & \textbf{92.8}            \\ 
		        \Xhline{2\arrayrulewidth}
			\end{tabular}
		}
	\end{center}
    \caption{\qy{Quantitative results on the ETH dataset. Note that, all methods are only trained on the indoor 3DMatch dataset. The FMR scores at $\tau_1$ = 10cm, $\tau_2$ = 5\% are compared.}}
	\label{tab:eth}
\end{table}

\textbf{Generalization from 3DMatch to ETH dataset.} \qy{Following the settings in~\cite{bai2020d3feat}, all models are only trained on the 3DMatch dataset, and then directly tested on the ETH dataset \cite{pomerleau2012challenging}. Note that, the ETH dataset consists of four scenes, \textit{i.e.}, Gazebo-Summer, Gazebo-Winter, Wood-Summer, and Wood-Autumn. Different from 3DMatch, the ETH dataset is acquired by static terrestrial scanners and dominated by outdoor vegetation, such as trees and bushes. In addition, the fragments of point clouds in the ETH dataset have lower resolution and contain more complex geometries compared with the 3DMatch dataset. The large domain gap between these two datasets poses a great challenge to the generalization of all approaches.}

\qy{
As shown in Table~\ref{tab:eth}, the performance of all baselines, namely D3Feat, FCGF, 3DMatch, and CGF, show a significant drop on the ETH dataset. Their FMR scores decrease up to 80\% compared with their results on the original 3DMatch dataset, as shown in Table \ref{tab:3dmatch}, and some techniques are even lower in performance than handcrafted descriptors such as SHOT. Fundamentally, the poor generalization of these methods is attributed to the fact that the descriptors learned by D3Feat, FCGF, and 3DMatch are variant to rigid transformations such as rotation and translation.} 

\begin{figure*}[thb]
\begin{center}
\includegraphics[width=0.97\linewidth]{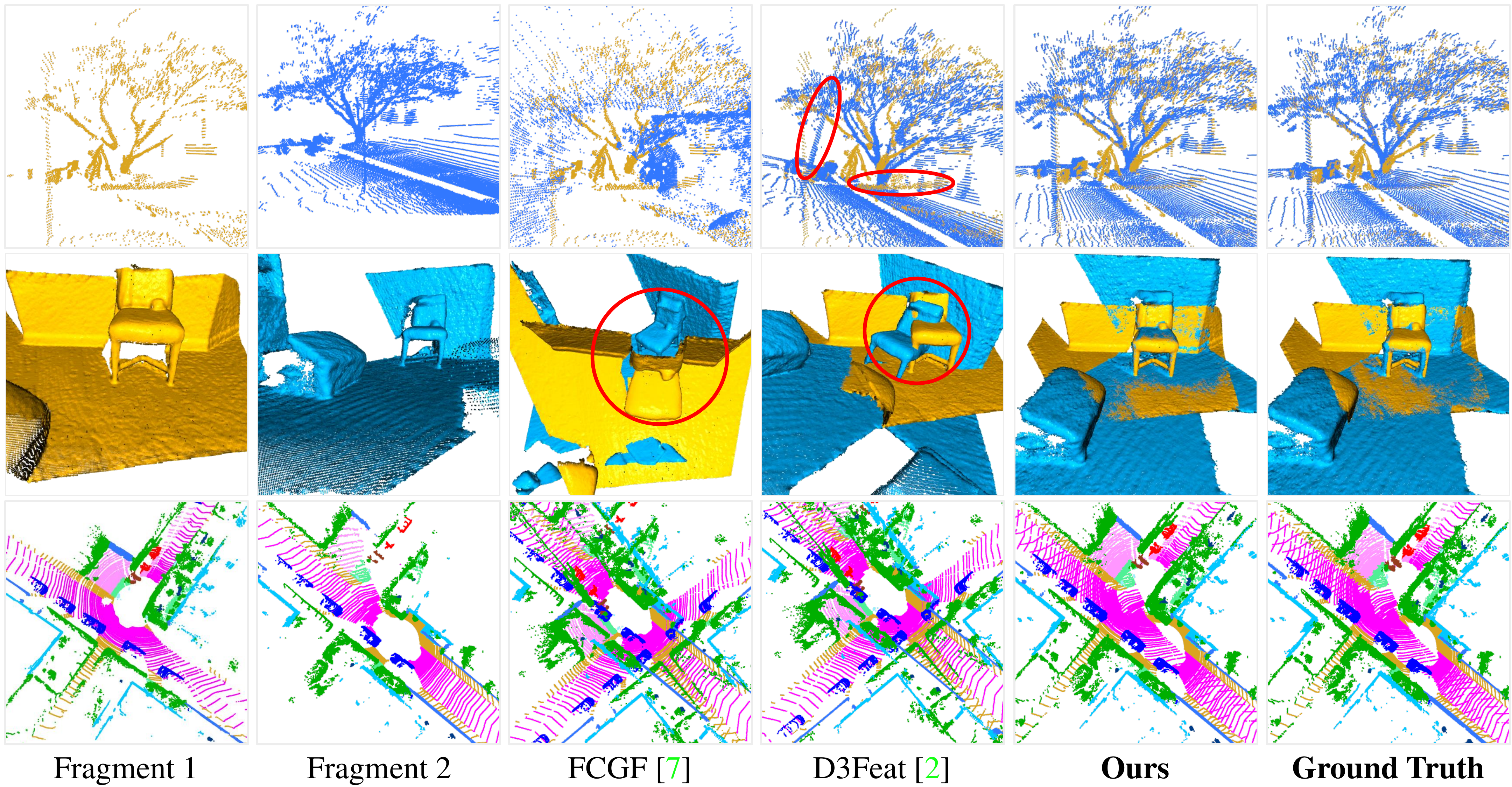}
\end{center}
\caption{\qy{Qualitative results of our method on unseen datasets. The first row is from 3DMatch to ETH, the second row is from KITTI to 3DMatch, and the third row is from 3DMatch to KITTI.}}
\label{fig:unseen_qualitative}
\end{figure*}

\bo{The descriptor generated by our \nickname{} achieves the highest FMR scores on all four scenes, significantly surpassing the second-best method (LMVD) by about 13\%. This clearly shows that our method has excellent generalization ability across the unseen dataset collected by a new sensor modality. This is primarily because our \nickname{} is explicitly designed to achieve rotational invariance. The first row of Figure \ref{fig:unseen_qualitative} shows the qualitative results.}

\begin{table}[thb]
	\begin{center}
		\scalebox{0.75}{
			\begin{tabular}{r|cc|cc}
             \Xhline{2\arrayrulewidth}
				\multirow{2}{*}{} & \multicolumn{2}{c|}{\bf{Origin}}  & \multicolumn{2}{c}{\bf{Rotated}} \\
				& FMR (\%)      & STD (\%)         & FMR (\%)      & STD (\%)   \\ \hline
				FPFH$^\dagger$~\cite{rusu2009fast}         & 35.9          & 13.4         & \underline{36.4}          & 13.6    \\
				SHOT$^\dagger$~\cite{tombari2010unique}         & 23.8          & 10.9         & 23.4          & 9.5     \\
				FCGF~\cite{choy2019fully}         & 32.5          & 7.4          & 1.0          & 1.0    \\
				D3Feat-rand~\cite{bai2020d3feat} & 60.7          & 7.7          & 17.2         &  4.6   \\
				D3Feat-pred~\cite{bai2020d3feat} & \underline{62.7}          & 8.1          & 17.8         &  3.2    \\
				\textbf{\nickname{} (Ours)}         & \textbf{84.5} & \textbf{5.9} & \textbf{84.2} & 5.8  \\ 
		        \Xhline{2\arrayrulewidth}
			\end{tabular}
		}
	\end{center}
	\caption{\qy{Quantitative results of different methods on the indoor 3DMatch dataset. Note that, all methods are only trained on the outdoor KITTI dataset.}}
	\label{tab:kitti_3dmatch}
\end{table}

\begin{table}[thb]
	\begin{center}
	\scalebox{0.8}{
	\begin{tabular}{r|cc|cc|c}
        \Xhline{2\arrayrulewidth}
		\multirow{2}{*}{} & \multicolumn{2}{c|}{\textbf{RTE (cm)}} & \multicolumn{2}{c|}{\textbf{RRE ($^\circ$)}} & \multirow{2}{*}{\textbf{Success (\%)}} \\
		& AVG                & STD               & AVG               & STD              &                       \\ \hline
		FCGF~\cite{choy2019fully}       & 27.1               & 5.58              & 1.61              & 1.51             & 24.19                 \\
		D3Feat-rand~\cite{bai2020d3feat}       & 37.8                   & 9.98                  & 1.58                  & 1.47                 & 18.47                      \\
		D3Feat-pred~\cite{bai2020d3feat}     & 31.6               & 10.1              & 1.44              & 1.35             & \underline{36.76}                 \\
		\textbf{\nickname{} (Ours)}      & \textbf{15.6}               & \textbf{1.89}                  & \textbf{0.98}                  & \textbf{0.63}                 & \textbf{81.44}                      \\ 		        \Xhline{2\arrayrulewidth}

	\end{tabular}
	}
	\end{center}
\caption{\qy{Quantitative results on the KITTI dataset. Note that, all models are trained on the indoor 3DMatch dataset, while being directly tested on the outdoor KITTI dataset. }}
  \label{tab:3dmatch_kitti}
\end{table}

\textbf{Generalization from KITTI to 3DMatch dataset.} \qy{All models are trained on the outdoor KITTI dataset which is mainly composed of sparse LiDAR point clouds, and then directly tested on the indoor 3DMatch dataset which consists of dense point clouds reconstructed from RGBD images. 
As presented in Table \ref{tab:kitti_3dmatch}, both D3Feat and FCGF achieve poor results on the 3DMatch dataset, especially when arbitary rotation in SO(3) exists. Their scores are even lower than the traditional methods such as FPFH, primarily because both D3Feat and FCGF have large numbers of parameters and tend to overfit the KITTI dataset, without learning the representative and general local patterns that can be applicable to the unseen dataset. By comparison, our \nickname{} achieves an overall FMR score of 79.6\%, demonstrating the superior generalization across novel scenarios. The second row of Figure \ref{fig:unseen_qualitative} shows the qualitative results. }

\textbf{Generalization from 3DMatch to KITTI dataset.} \qy{Additionaly, we evaluate the generalization ability from 3DMatch to KITTI dataset. All models are only trained on the indoor 3DMatch dataset, and then directly tested on the outdoor KITTI dataset. Table \ref{tab:3dmatch_kitti} presents the quantitative results. Because these two datasets are collected by different types of sensors, there is a large gap between the data distributions. Neither FCGF nor D3Feat can effectively generalize from 3DMatch to KITTI dataset. However, our method still demonstrates an excellent success rate of 69.19\%, doubling that of the second best method. The third row of Figure \ref{fig:unseen_qualitative} shows the qualitative results.}

\subsection{Ablation Study}
\label{Ex: abaltion}

\qy{
To systematically evaluate the effectiveness of each component in our \nickname{}, we conduct extensive ablative experiments on both 3DMatch and ETH datasets. In particular, we train all ablated models on the 3DMatch dataset, and then directly test them on both 3DMatch and ETH datasets.}

\smallskip\noindent\textbf{(1) Only removing the alignment with a reference axis.} \qy{
Initially, the reference axis is computed to align the input patch with the Z-axis. By removing this step, the rotation invariance on SO(3) is no longer maintained.}

\begin{table}[thb]
\begin{center}
	\scalebox{0.75}{
	\begin{tabular}{l|cc|cc|c|c}
        \Xhline{2\arrayrulewidth}
		& \multicolumn{4}{c|}{\bf{3DMatch}}                       & \multicolumn{2}{c}{\bf{ETH}} \\ \hline
		& \multicolumn{2}{c|}{\bf{Origin}} & \multicolumn{2}{c|}{\bf{Rotated}} 
		& {\bf{Origin}} 
		& {\bf{Rotated}} \\
		 \hfill Inlier ratio $\tau_2$ =  & 0.05         & 0.2         & 0.05          & 0.2         & 0.05  & 0.05 \\ \hline
		(1) W/o reference axis          & 95.1             & 80.0             & 63.0              & 23.1             & 83.9   & 13.5  \\
		(2) W/o transformation   & 93.5             & 70.8             & 87.7              &  44.8            & 60.5  & 47.7    \\
		(3) W/o Point Nets       & 94.0             &  66.3            &  93.8        &  66.1           & 42.6  & 42.4  \\
		(4) Replacing 3DCCN       & 64.4             & 10.6             & 64.4              & 10.4             & 0.0  & 0.0   \\
		(5) \textbf{The full method}     & \textbf{97.6}             & \textbf{85.7}             & \textbf{97.5}              & \textbf{86.1}             & \textbf{92.8}  &  \textbf{92.4}   \\  
        \Xhline{2\arrayrulewidth}
	\end{tabular}
	}
\end{center}
\caption{\qy{The FMR scores of all ablated models on the 3DMatch and ETH datasets with $\tau_1$ = 0.1cm.}}
\label{tab:ablation}
\end{table}

\smallskip\noindent\textbf{(2) Only removing the transformation on the XY-plane.} \qy{The transformation employed on the XY-plane is originally designed to eliminate the rotation variance of each voxel in the plane. In this experiment, we remove the transformation and directly operate on the non-transformed spherical voxels to formulate the cylindrical volume.}

\smallskip\noindent\textbf{(3) Only replacing the point-based layers with density.} \qy{ Instead of using the Point-based layers to learn a signature for each cylindrical voxel, we manually compute the point density of each voxel as its signature. Basically, this is to validate whether our point-based learned features are more general and representative than the commonly used, yet limited, handcrafted feature. }

\smallskip\noindent\textbf{(4) Only replacing 3DCCN by MLPs.} \qy{ The 3DCNN is designed to learn larger spatial structures from multiple voxels, whilst maintaining rotation equivariance. In this experiment, we replace the 3DCNN layers with the same number of MLP layers shared by all cylindrical voxels. These MLPs are unable to learn a wide context.
}

\textbf{Analysis.} \qy{Table \ref{tab:ablation} shows the results of all ablated networks on the 3DMatch dataset, as well as the generalization performance on the ETH datasets. It can be seen that: 1) Without using the alignment of a reference axis or the transformation of spherical voxels, the ablated models are unable to effectively match the point clouds either in 3DMatch or ETH datasets, especially for the point clouds with random rotations. This shows that the proposed Spatial Point Transformer indeed plays an important role to achieve rotation invariance in our \nickname{}.
2) Without using the advanced point-based neural layers to learn the signatures for spherical voxels, the ablated method can obtain consistent results on the 3DMatch dataset using the simple handcrafted feature, \textit{i.e.,} point density, but fails to generalize to the unseen ETH dataset. This clearly demonstrates that the learned local features tend to be much more powerful and general than the handcrafted features.
3) Without using the 3DCCN to learn larger surface structures, the ablated model only obtains significantly lower scores on both the 3DMatch and ETH datasets. This demonstrates that our 3DCCN is a key to preserving the local spatial patterns.
}

\section{Conclusion}
\bo{In this paper, we present a new neural descriptor to learn compact representations for complex 3D surfaces. The learned representations are rotation invariant, descriptive, and able to preserve complex local geometric patterns. Extensive experiments demonstrate that our descriptor has remarkable generalization ability across unseen scenarios and achieves superior results for 3D point cloud registration. \unclear{In future, we will investigate the integration  of keypoint detector, as well as the fully-convolutional architecture.
}
}

\clearpage

{\small
\bibliographystyle{ieee_fullname}
\bibliography{egbib}

\begin{thebibliography}{10}\itemsep=-1pt

\bibitem{ao2020repeatable}
Sheng Ao, Yulan Guo, Jindong Tian, Yong Tian, and Dong Li.
\newblock {A Repeatable and Robust Local Reference Frame for 3D Surface
  Matching}.
\newblock {\em PR}, 2020.

\bibitem{bai2020d3feat}
Xuyang Bai, Zixin Luo, Lei Zhou, Hongbo Fu, Long Quan, and Chiew-Lan Tai.
\newblock {D3Feat: Joint Learning of Dense Detection and Description of 3D
  Local Features}.
\newblock In {\em CVPR}, 2020.

\bibitem{Chen2019}
Chao Chen, Guanbin Li, Ruijia Xu, Tianshui Chen, Meng Wang, and Liang Lin.
\newblock {ClusterNet: Deep Hierarchical Cluster Network with Rigorously
  Rotation-Invariant Representation for Point Cloud Analysis}.
\newblock In {\em CVPR}, 2019.

\bibitem{chen20073d}
Hui Chen and Bir Bhanu.
\newblock {3D Free-form Object Recognition in Range Images Using Local Surface
  Patches}.
\newblock {\em PRL}, 2007.

\bibitem{choi2015robust}
Sungjoon Choi, Qian-Yi Zhou, and Vladlen Koltun.
\newblock {Robust Reconstruction of Indoor Scenes}.
\newblock In {\em CVPR}, 2015.

\bibitem{Choy2020}
Christopher Choy, Wei Dong, and Vladlen Koltun.
\newblock {Deep Global Registration}.
\newblock In {\em CVPR}, 2020.

\bibitem{choy20194d}
Christopher Choy, JunYoung Gwak, and Silvio Savarese.
\newblock {4D Spatio-Temporal Convnets: Minkowski Convolutional Neural
  Networks}.
\newblock In {\em CVPR}, 2019.

\bibitem{choy2019fully}
Christopher Choy, Jaesik Park, and Vladlen Koltun.
\newblock {Fully Convolutional Geometric Features}.
\newblock In {\em ICCV}, 2019.

\bibitem{chua1997point}
Chin~Seng Chua and Ray Jarvis.
\newblock {Point Signatures: A New Representation for 3D Object Recognition}.
\newblock {\em IJCV}, 1997.

\bibitem{Cohen2018}
Taco~S. Cohen, Mario Geiger, Jonas Koehler, and Max Welling.
\newblock {Spherical CNNs}.
\newblock In {\em ICLR}, 2018.

\bibitem{dai2017bundlefusion}
Angela Dai, Matthias Nie{\ss}ner, Michael Zollh{\"o}fer, Shahram Izadi, and
  Christian Theobalt.
\newblock {BundleFusion: Real-Time Globally Consistent 3D Reconstruction Using
  On-the-Fly Surface Reintegration}.
\newblock {\em TOG}, 2017.

\bibitem{Defferrard2020}
Michaël Defferrard, Martino Milani, Frédérick Gusset, and Nathanaël
  Perraudin.
\newblock {DeepSphere: A Graph-based Spherical CNN}.
\newblock In {\em ICLR}, 2020.

\bibitem{Deng2018a}
Haowen Deng, Tolga Birdal, and Slobodan Ilic.
\newblock {PPF-FoldNet: Unsupervised Learning of Rotation Invariant 3D Local
  Descriptors}.
\newblock In {\em ECCV}, 2018.

\bibitem{Deng2018}
Haowen Deng, Tolga Birdal, and Slobodan Ilic.
\newblock {PPFNet: Global Context Aware Local Features for Robust 3D Point
  Matching}.
\newblock In {\em CVPR}, 2018.

\bibitem{drost2010model}
Bertram Drost, Markus Ulrich, Nassir Navab, and Slobodan Ilic.
\newblock {Model Globally, Match Locally: Efficient and Robust 3D Object
  Recognition}.
\newblock In {\em CVPR}, 2010.

\bibitem{Elbaz2017}
Gil Elbaz, Tamar Avraham, and Anath Fischer.
\newblock {3D Point Cloud Registration for Localization using a Deep Neural
  Network Auto-Encoder}.
\newblock In {\em CVPR}, 2017.

\bibitem{Esteves2018}
Carlos Esteves, Christine Allen-Blanchette, Ameesh Makadia, and Kostas
  Daniilidis.
\newblock {Learning SO(3) Equivariant Representations with Spherical CNNs}.
\newblock In {\em ECCV}, 2018.

\bibitem{geiger2012we}
Andreas Geiger, Philip Lenz, and Raquel Urtasun.
\newblock {Are We Ready for Autonomous Driving? The KITTI Vision Benchmark
  Suite}.
\newblock In {\em CVPR}, 2012.

\bibitem{Gojcic2020a}
Zan Gojcic, Caifa Zhou, Jan~D. Wegner, Leonidas~J. Guibas, and Tolga Birdal.
\newblock {Learning Multiview 3D Point Cloud Registration}.
\newblock In {\em CVPR}, 2020.

\bibitem{Gojcic2019}
Zan Gojcic, Caifa Zhou, Jan~D Wegner, and Andreas Wieser.
\newblock {The Perfect Match: 3D Point Cloud Matching with Smoothed Densities}.
\newblock In {\em CVPR}, 2019.

\bibitem{sparseconv}
Benjamin Graham, Martin Engelcke, and Laurens Van Der~Maaten.
\newblock {3D Semantic Segmentation with Submanifold Sparse Convolutional
  Networks}.
\newblock In {\em CVPR}, 2018.

\bibitem{guo20143d}
Yulan Guo, Mohammed Bennamoun, Ferdous Sohel, Min Lu, and Jianwei Wan.
\newblock {3D Object Recognition in Cluttered Scenes with Local Surface
  Features: A Survey}.
\newblock {\em IEEE TPAMI}, 2014.

\bibitem{guo2016comprehensive}
Yulan Guo, Mohammed Bennamoun, Ferdous Sohel, Min Lu, Jianwei Wan, and
  Ngai~Ming Kwok.
\newblock {A Comprehensive Performance Evaluation of 3D Local Feature
  Descriptors}.
\newblock {\em IJCV}, 2016.

\bibitem{guo2013rotational}
Yulan Guo, Ferdous Sohel, Mohammed Bennamoun, Min Lu, and Jianwei Wan.
\newblock {Rotational Projection Statistics for 3D Local Surface Description
  and Object Recognition}.
\newblock {\em IJCV}, 2013.

\bibitem{guo2020deep}
Yulan Guo, Hanyun Wang, Qingyong Hu, Hao Liu, Li Liu, and Mohammed Bennamoun.
\newblock {Deep Learning for 3D Point Clouds: A Survey}.
\newblock {\em IEEE TPAMI}, 2020.

\bibitem{hu2020randla}
Qingyong Hu, Bo Yang, Linhai Xie, Stefano Rosa, Yulan Guo, Zhihua Wang, Niki
  Trigoni, and Andrew Markham.
\newblock {RandLA-Net: Efficient Semantic Segmentation of Large-Scale Point
  Clouds}.
\newblock In {\em CVPR}, 2020.

\bibitem{huang2020feature}
Xiaoshui Huang, Guofeng Mei, and Jian Zhang.
\newblock {Feature-metric Registration: A Fast Semi-supervised Approach for
  Robust Point Cloud Registration without Correspondences}.
\newblock In {\em CVPR}, 2020.

\bibitem{johnson1999using}
Andrew~E. Johnson and Martial Hebert.
\newblock {Using Spin Images for Efficient Object Recognition in Cluttered 3D
  Scenes}.
\newblock {\em IEEE TPAMI}, 1999.

\bibitem{joung2020cylindrical}
Sunghun Joung, Seungryong Kim, Hanjae Kim, Minsu Kim, Ig-Jae Kim, Junghyun Cho,
  and Kwanghoon Sohn.
\newblock {Cylindrical Convolutional Networks for Joint Object Detection and
  Viewpoint Estimation}.
\newblock In {\em CVPR}, 2020.

\bibitem{Khoury2017}
Marc Khoury, Qian-Yi Zhou, and Vladlen Koltun.
\newblock {Learning Compact Geometric Features}.
\newblock In {\em ICCV}, 2017.

\bibitem{Kim2020a}
Seohyun Kim, Jaeyoo Park, and Bohyung Han.
\newblock {Rotation-Invariant Local-to-Global Representation Learning for 3D
  Point Cloud}.
\newblock In {\em NeurIPS}, 2020.

\bibitem{kim2014shape2pose}
Vladimir~G Kim, Siddhartha Chaudhuri, Leonidas Guibas, and Thomas Funkhouser.
\newblock {Shape2pose: Human-Centric Shape Analysis}.
\newblock {\em TOG}, 2014.

\bibitem{kingma2014adam}
Diederik~P Kingma and Jimmy Ba.
\newblock {Adam: A Method for Stochastic Optimization}.
\newblock In {\em ICLR}, 2015.

\bibitem{li20073d}
Hongdong Li and Richard Hartley.
\newblock {The 3D-3D Registration Problem Revisited}.
\newblock In {\em ICCV}, 2007.

\bibitem{li2020end}
Lei Li, Siyu Zhu, Hongbo Fu, Ping Tan, and Chiew-Lan Tai.
\newblock {End-to-End Learning Local Multi-view Descriptors for 3D Point
  Clouds}.
\newblock In {\em CVPR}, 2020.

\bibitem{Lu2019}
Weixin Lu, Guowei Wan, Yao Zhou, Xiangyu Fu, Pengfei Yuan, and Shiyu Song.
\newblock {DeepVCP: An End-to-End Deep Neural Network for Point Cloud
  Registration}.
\newblock In {\em ICCV}, 2019.

\bibitem{ma2016fast}
Yanxin Ma, Yulan Guo, Jian Zhao, Min Lu, Jun Zhang, and Jianwei Wan.
\newblock {Fast and Accurate Registration of Structured Point Clouds with Small
  Overlaps}.
\newblock In {\em CVPRW}, 2016.

\bibitem{mian2010repeatability}
Ajmal Mian, Mohammed Bennamoun, and Robyn Owens.
\newblock {On the Repeatability and Quality of Keypoints for Local
  Feature-based 3D Object Retrieval from Cluttered Scenes}.
\newblock {\em IJCV}, 2010.

\bibitem{mian2006three}
Ajmal~S Mian, Mohammed Bennamoun, and Robyn Owens.
\newblock {Three-Dimensional Model-Based Object Recognition and Segmentation in
  Cluttered Scenes}.
\newblock {\em IEEE TPAMI}, 2006.

\bibitem{mishchuk2017working}
Anastasiia Mishchuk, Dmytro Mishkin, Filip Radenovic, and Jiri Matas.
\newblock {Working Hard to Know Your Neighbor's Margins: Local Descriptor
  Learning Loss}.
\newblock In {\em NeurIPS}, 2017.

\bibitem{novatnack2008scale}
John Novatnack and Ko Nishino.
\newblock {Scale-Dependent/Invariant Local 3D Shape Descriptors for Fully
  Automatic Registration of Multiple Sets of Range Images}.
\newblock In {\em ECCV}, 2008.

\bibitem{petrelli2011repeatability}
Alioscia Petrelli and Luigi Di~Stefano.
\newblock {On the Repeatability of The Local Reference Frame for Partial Shape
  Matching}.
\newblock In {\em ICCV}, 2011.

\bibitem{Poiesi2021}
Fabio Poiesi and Davide Boscaini.
\newblock {Distinctive 3D Local Deep Descriptors}.
\newblock In {\em ICPR}, 2021.

\bibitem{pomerleau2012challenging}
Fran{\c{c}}ois Pomerleau, Ming Liu, Francis Colas, and Roland Siegwart.
\newblock {Challenging Data Sets for Point Cloud Registration Algorithms}.
\newblock {\em IJRR}, 2012.

\bibitem{Rao2019}
Yongming Rao, Jiwen Lu, and Jie Zhou.
\newblock {Spherical Fractal Convolutional Neural Networks for Point Cloud
  Recognition}.
\newblock In {\em CVPR}, 2019.

\bibitem{rusu2009fast}
Radu~Bogdan Rusu, Nico Blodow, and Michael Beetz.
\newblock {Fast Point Feature Histograms (FPFH) for 3D Registration}.
\newblock In {\em ICRA}, 2009.

\bibitem{shotton2013scene}
Jamie Shotton, Ben Glocker, Christopher Zach, Shahram Izadi, Antonio Criminisi,
  and Andrew Fitzgibbon.
\newblock {Scene Coordinate Regression Forests for Camera Relocalization in
  RGB-D Images}.
\newblock In {\em CVPR}, 2013.

\bibitem{Spezialetti2019a}
Riccardo Spezialetti, Samuele Salti, and Luigi~Di Stefano.
\newblock {Learning an Effective Equivariant 3D Descriptor Without
  Supervision}.
\newblock In {\em ICCV}, 2019.

\bibitem{Spezialetti2020a}
Riccardo Spezialetti, Federico Stella, Marlon Marcon, Luciano Silva, Samuele
  Salti, and Luigi Di~Stefano.
\newblock {Learning to Orient Surfaces by Self-supervised Spherical CNNs}.
\newblock In {\em NeurlPS}, 2020.

\bibitem{stein1992structural}
Fridtjof Stein, G{\'e}rard Medioni, et~al.
\newblock {Structural Indexing: Efficient 3D Object Recognition}.
\newblock {\em IEEE TPAMI}, 1992.

\bibitem{kpconv}
Hugues Thomas, Charles~R Qi, Jean-Emmanuel Deschaud, Beatriz Marcotegui,
  Fran{\c{c}}ois Goulette, and Leonidas~J Guibas.
\newblock {Kpconv: Flexible and deformable convolution for point clouds}.
\newblock In {\em ICCV}, 2019.

\bibitem{tombari2010unique}
Federico Tombari, Samuele Salti, and Luigi Di~Stefano.
\newblock {Unique Signatures of Histograms for Local Surface Description}.
\newblock In {\em ECCV}, 2010.

\bibitem{Tombari2010shot}
Federico Tombari, Samuele Salti, and Luigi Di~Stefano.
\newblock {Unique Signatures of Histograms for Local Surface Description}.
\newblock In {\em ECCV}, 2010.

\bibitem{valentin2016learning}
Julien Valentin, Angela Dai, Matthias Nie{\ss}ner, Pushmeet Kohli, Philip Torr,
  Shahram Izadi, and Cem Keskin.
\newblock {Learning to Navigate the Energy Landscape}.
\newblock In {\em 3DV}, 2016.

\bibitem{xiao2013sun3d}
Jianxiong Xiao, Andrew Owens, and Antonio Torralba.
\newblock {Sun3d: A Database of Big Spaces Reconstructed using SfM and Object
  Labels}.
\newblock In {\em ICCV}, 2013.

\bibitem{yang2019learning}
Bo Yang, Jianan Wang, Ronald Clark, Qingyong Hu, Sen Wang, Andrew Markham, and
  Niki Trigoni.
\newblock {Learning Object Bounding Boxes for 3D Instance Segmentation on Point
  Clouds}.
\newblock In {\em NeurIPS}, 2019.

\bibitem{Yang2018}
Yaoqing Yang, Chen Feng, Yiru Shen, and Dong Tian.
\newblock {FoldingNet: Point Cloud Auto-encoder via Deep Grid Deformation}.
\newblock In {\em CVPR}, 2018.

\bibitem{Yew2018}
Zi~Jian Yew and Gim~Hee Lee.
\newblock {3DFeat-Net: Weakly Supervised Local 3D Features for Point Cloud
  Registration}.
\newblock In {\em ECCV}, 2018.

\bibitem{yew2020rpm}
Zi~Jian Yew and Gim~Hee Lee.
\newblock {RPM-Net: Robust Point Matching using Learned Features}.
\newblock In {\em CVPR}, 2020.

\bibitem{you2020pointwise}
Yang You, Yujing Lou, Qi Liu, Yu-Wing Tai, Lizhuang Ma, Cewu Lu, and Weiming
  Wang.
\newblock {Pointwise Rotation-Invariant Network with Adaptive Sampling and 3D
  Spherical Voxel Convolution}.
\newblock In {\em AAAI}, 2020.

\bibitem{Yu2020}
Ruixuan Yu, Xin Wei, Federico Tombari, and Jian Sun.
\newblock {Deep Positional and Relational Feature Learning for
  Rotation-Invariant Point Cloud Analysis}.
\newblock In {\em ECCV}, 2020.

\bibitem{zaharescu2009surface}
Andrei Zaharescu, Edmond Boyer, Kiran Varanasi, and Radu Horaud.
\newblock {Surface Feature Detection and Description with Applications to Mesh
  Matching}.
\newblock In {\em ICCV}, 2009.

\bibitem{Zeng2017}
Andy Zeng, Shuran Song, Matthias Nie{\ss}ner, Matthew Fisher, Jianxiong Xiao,
  and Thomas Funkhouser.
\newblock {3DMatch: Learning Local Geometric Descriptors from RGB-D
  Reconstructions}.
\newblock In {\em CVPR}, 2017.

\bibitem{Zhao2020}
Yongheng Zhao, Tolga Birdal, Jan~Eric Lenssen, Emanuele Menegatti, Leonidas
  Guibas, and Federico Tombari.
\newblock {Quaternion Equivariant Capsule Networks for 3D Point Clouds}.
\newblock In {\em ECCV}, 2020.

\bibitem{Zhong2010Intrinsic}
Yu Zhong.
\newblock {Intrinsic Shape Signatures: A Shape Descriptor for 3D Object
  Recognition}.
\newblock In {\em ICCVW}, 2009.

\bibitem{Zhou2018}
Lei Zhou, Siyu Zhu, Zixin Luo, Tianwei Shen, Runze Zhang, Mingmin Zhen, Tian
  Fang, and Long Quan.
\newblock {Learning and Matching Multi-view Descriptors for Registration of
  Point Clouds}.
\newblock In {\em ECCV}, 2018.

\end{thebibliography}
}

\newpage
\appendix
\section*{Appendix} 
\section{Definitions of Equivariance and Invariance}
\qysupp{For a specified function $f: X \rightarrow Y$ as well as a specified group action $G$, $f$ is said to be equivariant with respect to transformation action $g \in G$ if,}
\begin{equation} \label{Supp-eq1} 
\begin{split}
f(g \circ x) = g \circ f(x), \quad x \in X
\end{split}
\end{equation}
\qysupp{Analogously, $f$ is said to be invariant to transformations $g \in G$ when the following equation is satisfied: }
\begin{equation} \label{Supp-eq2} 
\begin{split}
f(g \circ x) = f(x), \quad x \in X
\end{split}
\end{equation}

\section{Theoretical Proof of Equivariance}

\paragraph{Lemma 1.}\textit{Given a discrete 2D rotation group\footnote{The minimum rotation unit depends on the way partion along the azimuth axis. \textit{i.e.,} $\frac{2\pi}{L}$.} $\mathcal{R} \subset \mathbf{SO}\rm{(2)}$, where $\mathcal{R}=\{r_i \in \mathbb{R}^{3 \times 3}, i = 1,2,...,L\}$, then the proposed spatial point transformer is an equivariant map for the 2D rotation group $\mathcal{R}$.}

\paragraph{Proof:} For a local patch $\mathbf{P}^\mathrm{s}$, the spatial point transformer in our framework can be regarded as a mapping $\mathcal{M}_v$ from $\mathbf{P}^\mathrm{s}$ to cylindrical volume $\mathbf{C} \in \mathbb{R}^{J \times K \times L \times k_v \times 3}: \mathbb{R}^{3 \times |\mathbf{P}^\mathrm{s}|} \rightarrow \mathbb{R}^{J \times K \times L \times k_v \times 3}$. For a group action $r_i$ in $\mathcal{R}$, suppose $\mathbf{\widetilde{P}}^\mathrm{s} = r_i \circ \mathbf{P}^\mathrm{s} = r_i \mathbf{P}^\mathrm{s}$, and the rotated local neighbouring set  $\mathbf{\widetilde{P}}_{jk(l+i)}=r_i \mathbf{P}_{jkl}$. On the other hand, for the rotation matrix defined in Eq.~\ref{eq3}, we have $\mathbf{R}_{jkl}=\mathbf{R}_{jk(l+i)}   1 12 r_i$. Then, the $(j^{th}, k^{th}, l^{th})$ element $\mathbf{c}^p_{jkl}$ of cylindrical volume $\mathbf{C}$ satisfies:
\begin{equation} \label{Supp-eq3} 
\begin{split}
\mathbf{c}^p_{jkl} &= \mathbf{R}_{jkl} \mathbf{P}_{jkl} = \mathbf{R}_{jk(l+i)} r_i \mathbf{P}_{jkl} \\
                   &= \mathbf{R}_{jk(l+i)}\mathbf{\widetilde{P}}_{jk(l+i)}= \mathbf{c}^{\widetilde{p}}_{jk(l+i)},
\end{split}
\end{equation}
where $\mathbf{c}^{\widetilde{p}}_{jk(l+i)}\in \widetilde{\mathbf{C}}$, which is the cylindrical volume corresponding to the $\mathbf{\widetilde{P}}^\mathrm{s}$ . Based on Eq.~\ref{Supp-eq3}, we can infer that $\mathbf{c}^p_{jk(l-i)}=\mathbf{c}^{\widetilde{p}}_{jkl}$, hence the transformed cylindrical volume $\widetilde{\mathbf{C}}$ can be formulated as:
\begin{equation} \label{Supp-eq4}
\begin{split}
\widetilde{\mathbf{C}} = &\mathcal{M}_v(r_i \circ \mathbf{P}^\mathrm{s}) = \mathcal{M}_v(\mathbf{\widetilde{P}}^\mathrm{s})  \\
&= [\mathbf{c}^{\widetilde{p}}_{111},...,\mathbf{c}^{\widetilde{p}}_{jkl},...,\mathbf{c}^{\widetilde{p}}_{JKL}]\\
&= [\mathbf{c}^p_{11(1-i)},...,\mathbf{c}^p_{jk(l-i)},...,\mathbf{c}^p_{JK(L-i)}],
\end{split}
\end{equation}
where $\mathbf{c}^p_{jkl} = \mathbf{c}^p_{jk(l+L)}$ if $l<1$, due to the periodic property of the cylindrical volume in the XY plane. On the other hand, $r_i \circ \mathcal{M}_v$ means rotating the cylindrical volume $\mathbf{C}$ around the Z-axis, that is:
\begin{equation} \label{Supp-eq5}
\begin{split}
&r_i \circ \mathcal{M}_v(\mathbf{P}^\mathrm{s}) = r_i \circ [\mathbf{c}^p_{111},...,\mathbf{c}^p_{jkl},...,\mathbf{c}^p_{JKL}] \\
&= [\mathbf{c}^p_{11(1-i)},...,\mathbf{c}^p_{jk(l-i)},...,\mathbf{c}^p_{JK(L-i)}] \\
&= \mathcal{M}_v(r_i \circ \mathbf{P}^\mathrm{s}),
\end{split}
\end{equation}
which completes our proof that the spatial point transformer $\mathcal{M}_v$ is an equivariant map for the rotation group $\mathcal{R}$.

\paragraph{Lemma 2.} \textit{Given a discrete 2D rotation group $\mathcal{R} \subset \mathbf{SO}\rm{(2)}$, where $\mathcal{R}=\{r_i \in \mathbb{R}^{3 \times 3}, i = 1,2,...,L\}$, then 3DCCN is an equivariant map for the 2D rotation group $\mathcal{R}$.} 

\paragraph{Proof:} The proposed 3D cylindrical convolution can be formulated as a set of convolution filter $\psi^i$ on the cylindrical feature maps $f$: 

\begin{equation} \label{Supp-eq7}
\begin{split}
&(f\ast\psi^i)(\rho,z,\theta)= \\
&\sum_d\sum_{j}\sum_{k}\sum_{l} f_d(j,k,l) \psi^{i}_d(j-\rho,k-z,l-\theta),
\end{split}
\end{equation}
where $\rho$, $\theta$ and $z$ denote radial distance, azimuth angle and height, respectively. $d$ is the number of channels in feature map.

Suppose a group action $r_i$ in $\mathcal{R}$ operating on cylindrical feature maps $f$, we have $(r_i \circ f)(\rho,z,\theta) = f(\rho,z,\theta-i)$. To clarify,  the $r_i$-transformed feature maps $r_i \circ f$ at the coordinate $(\rho,z,\theta)$ is equivalent to find the value  in the original feature map $f$ at the coordinate $(\rho,z,\theta-i)$. Leaving out the summation over feature maps for clarity, we have:
\begin{equation} \label{Supp-eq8}
\begin{split}
&((r_i \circ f)\ast\psi^i)(\rho,z,\theta)= \\
&\sum_{j}\sum_{k}\sum_{l} f(j,k,l-i) \psi^{i}(j-\rho,k-z,l-\theta).
\end{split}
\end{equation}
Using the substitution $l \rightarrow l+i$, then Eq.~\ref{Supp-eq8} can be transformed into:
\begin{equation} \label{Supp-eq9}
\begin{split}
&((r_i \circ f)\ast\psi^i)(\rho,z,\theta) \\
&=\sum_{j}\sum_{k}\sum_{l} f(j,k,l) \psi^{i}(j-\rho,k-z,l-(\theta-i)) \\
&= (f\ast\psi^i)(\rho,z,\theta-i) \\
&= (r_i \circ (f\ast\psi^i))(\rho,z,\theta),
\end{split}
\end{equation}
which completes our proof that 3DCCN is an equivariant map for the 2D rotation group $\mathcal{R}$.

\begin{figure*}[thb]
	\begin{center}
		\includegraphics[width=1.0\linewidth]{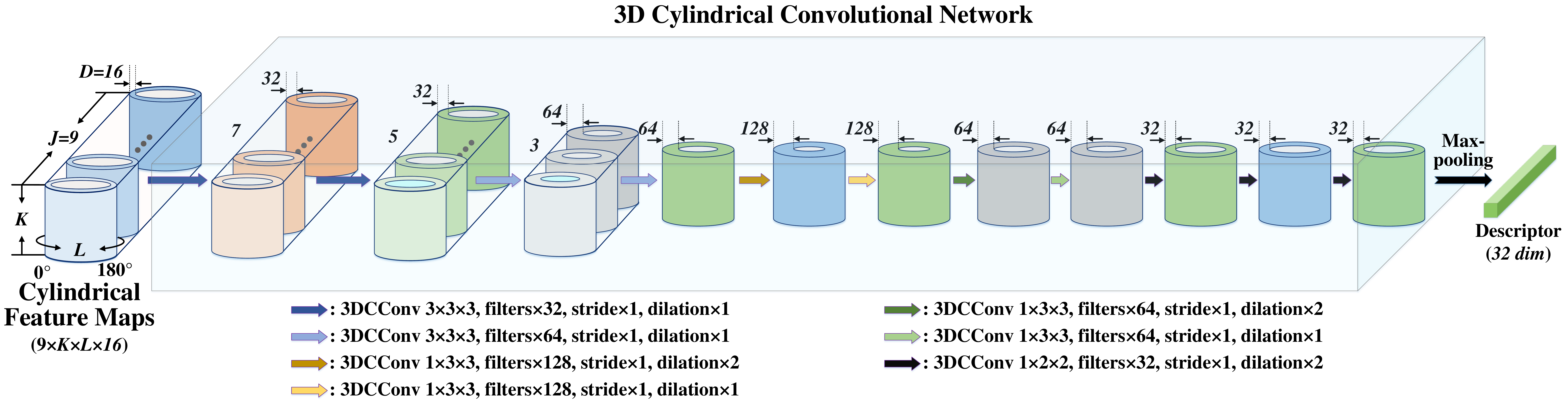}
	\end{center}
	\caption{Detailed architecture of our proposed 3D cylindrical convolution networks.}
	\label{fig:supp_network_architecture}
\end{figure*}

\section{Detailed Network Architecture}
\qysupp{Using 3D Cylindrical Convolution (3D-CCN) as a basic operator, we build a hierarchical learning architecture as depicted in Figure \ref{fig:supp_network_architecture}. To ensure the reproducibility of our framework, we also provide detailed information on the kernel size, stride, and the number of filters in this figure. A number of cylindrical convolution layers are stacked together to progressively learn descriptive, yet compact local feature representations. In particular, the maximum number of channels used in our cylindrical feature map is 128, which is much smaller than 1024 used in D3Feat \cite{bai2020d3feat}. This further makes our network very lightweight and less prone to overfitting.}

\section{Detailed Evaluation Metrics}
\qysupp{We further provide the detailed evaluation metrics used in our experiments (Sec.~\ref{Experiments}).}
\medskip\\
{\bf Evaluation Metrics on 3DMatch and ETH.}\quad \qysupp{We adopt Feature Matching Recall ($\mathrm{FMR}$) as the main evaluation metric to evaluate the performance of the learned descriptors. Similar to~\cite{choi2015robust,Deng2018,Deng2018a,choy2019fully}, we also provide a formal definition for each metric as follows.}

\qysupp{First, suppose there are a total of $H$ pairs of fragments in the 3DMatch dataset, where the overlap is greater than 30\%. Each pair of fragments $\mathcal{P}_h$ and $\mathcal{Q}_h$ can be aligned by the ground-truth rigid transformation $\mathbf{T}_h=\{ \mathbf{R}_h, \mathbf{t}_h \}$. Then, we randomly select $n$ points from the two point clouds to obtain $\mathcal{P}_h^n=\{ \mathbf{p}_1, \mathbf{p}_2,..., \mathbf{p}_n \}$ and $\mathcal{Q}_h^n=\{ \mathbf{q}_1, \mathbf{q}_2,..., \mathbf{q}_n\}$. In particular, a set of point correspondences $\mathbf{\Omega}_h$ between $\mathcal{P}_h^n$ and $\mathcal{Q}_h^n$ is also generated by applying nearest neighbor search $\mathrm{NN}$ in the feature space $\mathcal{M}$:
}

\qysupp{Then the average feature matching recall on the 3DMatch dataset is defined as:}
\begin{equation} \label{Supp-eq11}
\begin{split}
&\mathrm{FMR}= \\ 
& \frac{1}{H} \sum_{h=1}^{H} \mathds{1}\bigg(\Big[ \frac{1}{|\mathbf{\Omega}_h|} \sum_{(\mathbf{p}_i, \mathbf{q}_j) \in \mathbf{\Omega}_h}\mathds{1}(\left\| \mathbf{p}'_i - \mathbf{q}_j \right\| < \tau_1)\Big] > \tau_2\bigg),
\end{split}
\end{equation}

\noindent \qysupp{where $\mathbf{p}'_i = \mathbf{R}_h\mathbf{p}_i + \mathbf{t}_h$, $||\cdot||$ denotes the Euclidean distance, $\tau_1$ 
and $\tau_2$ is the inlier distance threshold and inliner ratio threshold, respectively. $\mathds{1}$ is the indicator function. $\mathbf{\Omega}_h$ denotes a set of point correspondences between $\mathcal{P}_h^n$ and $\mathcal{Q}_h^n$. In particular, it is generated by applying nearest neighbor search $\mathrm{NN}$ in the feature space $\mathcal{M}$:}
\begin{equation} \label{Supp-eq10}
\begin{split}
\mathbf{\Omega}_h = \{ \{\mathbf{p}_i, \mathbf{q}_j\} | &\mathcal{M}(\mathbf{p}_i)=\mathrm{NN}(\mathcal{M}(\mathbf{q}_j), \mathcal{M}(\mathcal{P}_h^n)), \\
  &\mathcal{M}(\mathbf{q}_j)=\mathrm{NN}(\mathcal{M}(\mathbf{p}_i), \mathcal{M}(\mathcal{Q}_h^n)) \}.
\end{split}
\end{equation}

\begin{comment}
\qysupp{Accordingly, inlier ratio $\mathrm{IR}$ is defined as:}
\begin{equation} \label{Supp-eq12}
\begin{split}
\mathrm{IR} = \frac{1}{H} \sum_{h=1}^{H}\Big[ \frac{1}{|\mathbf{\Omega}_h|} \sum_{(\mathbf{p}_i, \mathbf{q}_j) \in \mathbf{\Omega}_h}\mathds{1}(\left\| \mathbf{p}'_i - \mathbf{q}_j \right\| < \tau_1)\Big].
\end{split}
\end{equation}

\qysupp{Different from FMR and IR metrics, which are mainly used to evaluate the quality of matching pairs, the registration recall is introduced to evaluate the performance of descriptors in an entire registration pipeline. }
\begin{equation} \label{Supp-eq13}
\begin{split}
&\mathrm{RR} = \\
&\frac{1}{H} \sum_{h=1}^{H}\mathds{1}\bigg( \sqrt{\frac{1}{|\mathbf{\Omega}^*_h|} \sum_{(\mathbf{p}^*, \mathbf{q}^*) \in \mathbf{\Omega}^*_h} \left\| \mathbf{\hat{R}}_h\mathbf{p}^*\!+\!\mathbf{\hat{t}}_h\!-\! \mathbf{q}^* \right\|^2 } < \tau_3 \bigg),
\end{split}
\end{equation}
\qysupp{where $\mathbf{\Omega}^*_h = \{ \{\mathbf{p}^*,\mathbf{q}^*\} | \mathbf{p}^* \in \mathcal{P}_h, \mathbf{q}^* \in \mathcal{Q}_h \}$ is a set of  ground-truth corresponding points between $\mathcal{P}_h$ and $\mathcal{Q}_h$, and $\mathbf{\hat{T}}_h = \{ \mathbf{\hat{R}}_h,\mathbf{\hat{t}}_h \}$ is the rigid transformation matrix estimated by using RANSAC~\cite{fischler1981random}. $\tau_3$ is the Mean Squared Error (MSE) threshold. Following~\cite{bai2020d3feat}, we perform up to 50,000 iterations on the initial correspondence set in our experiment to estimate the rigid transformations, and set the MSE threshold to 0.2.}
\end{comment}

\noindent{\bf Evaluation Metrics on KITTI.}\quad \qysupp{Different from the indoor 3DMatch dataset, the evaluation metrics on the KITTI dataset are Relative Translational Error ($\mathrm{RTE}$), Relative Rotation Error ($\mathrm{RRE}$), and Success Rate ($\mathrm{SR}$), respectively. According to the definitions in~\cite{ma2016fast,Yew2018,choy2019fully}, for a pair of fragments $\mathcal{P}_h$ and $\mathcal{Q}_h$, the relative rotation error $\mathrm{RRE}$ is calculated as:}
\begin{equation} \label{Supp-eq14}
\begin{split}
\mathrm{RRE} = \arccos{\bigg( \frac{\text{trace}(\mathbf{\hat{R}}_h^T \mathbf{R}_h)-1}{2} \bigg)}\frac{180}{\pi},
\end{split}
\end{equation}
\qysupp{where $\mathbf{R}_h$ and $\mathbf{\hat{R}}_h$ denote the ground-truth and the estimated rotation matrix, respectively. Analogously, the relative translation error $\mathrm{RTE}$ can be calculated by:}
\begin{equation} \label{Supp-eq15}
\begin{split}
\mathrm{RTE} = \left\| \mathbf{\hat{t}}_h - \mathbf{t}_h \right\|,
\end{split}
\end{equation}
\qysupp{where $\mathbf{t}_h$ and $\mathbf{\hat{t}}_h$ denote the ground-truth and the estimated translation matrix, respectively. Finally, success rate $\mathrm{SR}$ is defined as:}
\begin{equation} \label{Supp-eq16}
\begin{split}
\mathrm{SR} = \frac{1}{H} \sum_{h=1}^{H}\mathds{1}\bigg( \mathrm{RRE}_h < 2\text{m}\ \&\&\ \mathrm{RTE}_h < 5^\circ \bigg).
\end{split}
\end{equation}

\begin{table*}[thb]
	\centering
	\resizebox{1.0\textwidth}{!}{%
		\begin{tabular}{lccccccccccc}
        \toprule[1.0pt]
			& FPFH~\cite{rusu2009fast}& SHOT~\cite{Tombari2010shot}& 3DMatch~\cite{Zeng2017}& CGF$^\dagger$~\cite{Khoury2017} & PPFNet~\cite{Deng2018}& PPF-FoldNet~\cite{Deng2018a}& PerfectMatch~\cite{Gojcic2019}  & FCGF~\cite{choy2019fully}& D3Feat~\cite{bai2020d3feat}& LMVD~\cite{li2020end} & Ours      \\
			\midrule               
			Kitchen & 30.6       & 17.8         & 57.5           & 46.1         & 89.7              & 78.7         & 97.0              & -   & -     & \textbf{99.4}   & \underline{99.2} \\
			Home 1  & 58.3        & 37.2        & 73.7          & 61.5       & 55.8        & 76.3    & 95.5           & -  &-    & \textbf{98.7}   & \underline{98.1} \\
			Home 2  & 46.6          & 33.7         & 70.7          & 56.3       & 59.1           & 61.5            & 89.4        & -   & -         & \underline{94.7}   & \textbf{96.2} \\
			Hotel 1 & 26.1     & 20.8        & 57.1     & 44.7       & 58.0         & 68.1               & \underline{96.5}  & -  & -        & \textbf{99.6}   & \textbf{99.6}\\
			Hotel 2 & 32.7       & 22.1        & 44.2           & 38.5       & 57.7        & 71.2   & 93.3        & -  & -       & \textbf{100.0}  & \underline{97.1} \\
			Hotel 3 & 50.0       & 38.9        & 63.0        & 59.3   & 61.1    & 94.4    & \underline{98.2} & -   & -     & \textbf{100.0}   & \textbf{100.0} \\
			Study   & 15.4       & 7.2        & 56.2          & 40.8      & 53.4     & 62.0     & 94.5  & - & -   & \underline{95.5} & \textbf{95.6}\\
			MIT Lab & 27.3   & 13.0   & 54.6    & 35.1   & 63.6    & 62.3    & 93.5    & -  & -    & \underline{92.2} & \textbf{94.8} \\
			\midrule          
			\textbf{Average} & 35.9  & 23.8     & 59.6   & 47.8        & 62.3    & 71.8  & 94.7  & 95.2  & 95.8  & \underline{97.5}  & \textbf{97.6} \\
			\textbf{STD} & 13.4  & 10.9     & 8.8   & 9.4  & 10.8  & 10.5    & \underline{2.7}  & 2.9  & 2.9  & 2.8  & \textbf{1.9} \\
        \toprule[1.0pt]
		\end{tabular}
	}
	\caption{Average recall (\%) {of different methods on the 3DMatch benchmark with $\tau_1 = 10$cm and $\tau_2 = 0.05$}. The symbol '-' means the results are unavailable and $\dagger$ means the results are reported from~\cite{Deng2018a}, which is different from Table~\ref{tab:3dmatch}. }
	\label{tab:supp_3dmatch_recall_origin}
\end{table*}

\begin{table*}[thb]
	\centering
	\resizebox{1.0\textwidth}{!}{%
		\begin{tabular}{lccccccccccc}
        \toprule[1.0pt]
			& FPFH~\cite{rusu2009fast}& SHOT~\cite{Tombari2010shot}& 3DMatch~\cite{Zeng2017}& CGF$^\dagger$~\cite{Khoury2017} & PPFNet~\cite{Deng2018}& PPF-FoldNet~\cite{Deng2018a}& PerfectMatch~\cite{Gojcic2019}  & FCGF~\cite{choy2019fully}& D3Feat~\cite{bai2020d3feat}& LMVD~\cite{li2020end} & Ours      \\
			\midrule               
			Kitchen & 29.1       & 17.8         & 0.4           & 44.7         & 0.2              & 78.9         & \underline{97.2}              & -   & -     & -   & \textbf{99.0} \\
			Home 1  & 59.0        & 35.6        & 1.3         & 66.7      & 0.0        & 78.2    & \underline{96.2}           & -  & -    & -   & \textbf{98.7} \\
			Home 2  & 47.1          & 33.7         & 3.4          & 52.9       & 1.4           & 64.4            & \underline{90.9}        & -   & -         & -   & \textbf{96.2} \\
			Hotel 1 & 30.1     & 21.7        & 0.4     & 44.3       & 0.4         & 67.7               & \underline{96.5}  & -  & -        & -  & \textbf{99.6}\\
			Hotel 2 & 30.0       & 24.0        & 0.0          & 44.2       & 0.0       & 62.9   & \underline{92.3}        & -  & -       & -  & \textbf{97.1} \\
			Hotel 3 & 51.9       & 33.3        & 1.0        & 63.0   & 0.0    & 96.3    & \underline{98.2} & -   & -     & -   & \textbf{100.0} \\
			Study   & 15.8       & 8.2        & 0.0         & 41.8      & 0.0     & 62.7     & \underline{94.5}  & - & -   & - & \textbf{94.9}\\
			MIT Lab & 41.6   & 62.3   & 3.9    & 45.5              & 0.0    & 67.5    & \underline{93.5}    & -  & -    & - & \textbf{94.8} \\
			\midrule          
			\textbf{Average} & 36.4  & 23.4     & 1.1   & 49.9        & 0.3    & 73.1  & 94.9  & 95.3  & 95.5  & \underline{96.9}  & \textbf{97.5} \\
			\textbf{STD} & 13.6  & 9.5     & 1.2   & 8.9  & 0.5  & 10.4    & 2.5  & 3.3  & 3.5  & -  & 1.9 \\
        \toprule[1.0pt]
		\end{tabular}
	}
	\caption{Average recall (\%) {of different methods on the rotated 3DMatch benchmark with $\tau_1 = 10$cm and $\tau_2 = 0.05$}. The symbol '-' means the results are unavailable and $\dagger$ means the results are reported from~\cite{Deng2018a}, which is different from Table~\ref{tab:3dmatch}. }
	\label{tab:supp_3dmatch_recall_rotate}
\end{table*}

\section{Implementation Details}
\qysupp{Here we provide extra implementation details in this section. The detailed hyperparameter settings of our \nickname{} on different datasets are listed in Table \ref{tab:hyperparameter}. In particular, we keep the same parameter settings as the training dataset when generalized to unseen datasets, except for the support radius $R$ and query radius $R_v$, due to the varying point densities in different datasets. Specifically, we follow the scheme in D3Feat \cite{bai2020d3feat} to adaptively adjust the radius according to the ratio.
}
% Dataset preprocessing details
\begin{table}[thb]
\begin{center}
\scalebox{0.9}{
\begin{tabular}{rcccccc}
\toprule[1.0pt]
Dataset & $J$ & $K$  & $L$  & $R$  & $R_v$  & $k_v$ \\
\midrule[0.75pt]
3DMatch \cite{Zeng2017} & 9 & 40 & 80 & 0.3m & 0.04m & 30   \\
KITTI \cite{geiger2012we}   & 9 & 30 & 60 & 2.0m & 0.30m & 30   \\
ETH \cite{pomerleau2012challenging} & 9 & 40 & 80 & 0.8m & 0.10m & 30   \\
\toprule[1.0pt]
\end{tabular}
}
\end{center}
\caption{The hyperparameters set by our method in different datasets.}
\label{tab:hyperparameter}
\end{table}

\section{Additional Results on 3DMatch}

\qysupp{For comparison, we also report the detailed quantitative results of our \nickname{} on the 3DMatch dataset in Table \ref{tab:supp_3dmatch_recall_origin} and the rotated 3DMatch dataset in Table \ref{tab:supp_3dmatch_recall_rotate}.}

\section{Additional qualitative results.}
\qysupp{As illustrated in Sec. \ref{Ex:generalization}, our \nickname{} has demonstrated  superior quantitative generalization performance across different datasets with different sensor modalities. Here, we further show additional qualitative results in this section.}

\begin{figure*}[thb]
	\begin{center}
		\includegraphics[width=1.0\linewidth]{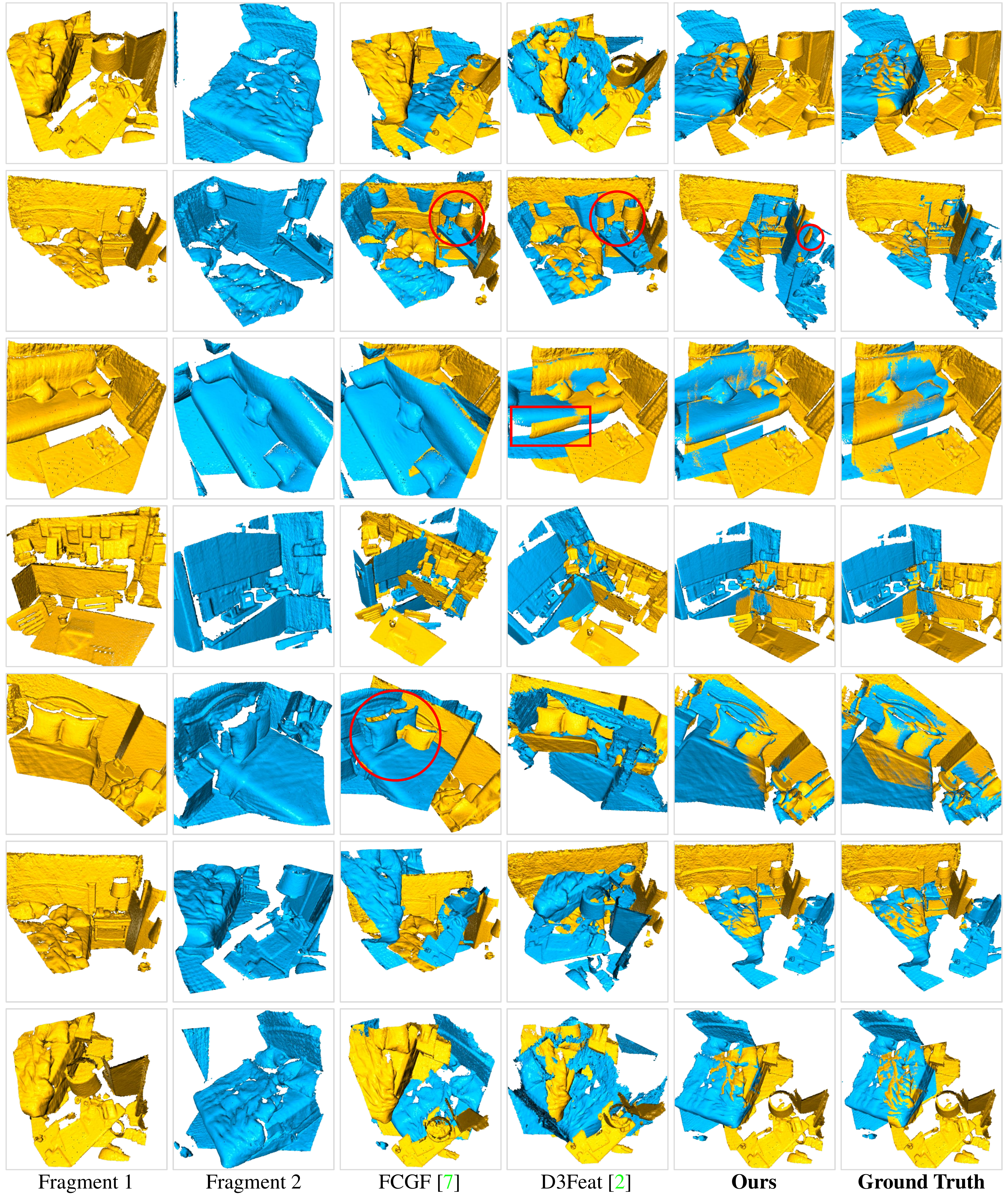}
	\end{center}
	\caption{\qysupp{Additional qualitative results achieved by FCGF \cite{choy2019fully}, D3Feat \cite{bai2020d3feat}, and our \textbf{\nickname{}} on the 3DMatch dataset. Note that, all methods are only trained on the outdoor KITTI \cite{geiger2012we} dataset. Red boxes/circles show the failure cases.}}
	\label{fig:supp_qualitative_3DMatch}
\end{figure*}

\smallskip\noindent\textbf{Additional qualitative results on the 3DMatch dataset.} \qysupp{We first show the additional qualitative results achieved by FCGF \cite{choy2019fully}, D3Feat \cite{bai2020d3feat}, and our \nickname{} on the 3DMatch dataset in Fig. \ref{fig:supp_qualitative_3DMatch}. It can be seen that the FCGF and D3Feat are prone to mismatching the fragments when the two input partial scans have relatively significant differences. However, our simple \nickname{} can always achieve consistent registration performance on this dataset, despite only being trained on the outdoor KITTI dataset with sparse LiDAR point clouds.}

\begin{figure*}[thb]
	\begin{center}
		\includegraphics[width=1.0\linewidth]{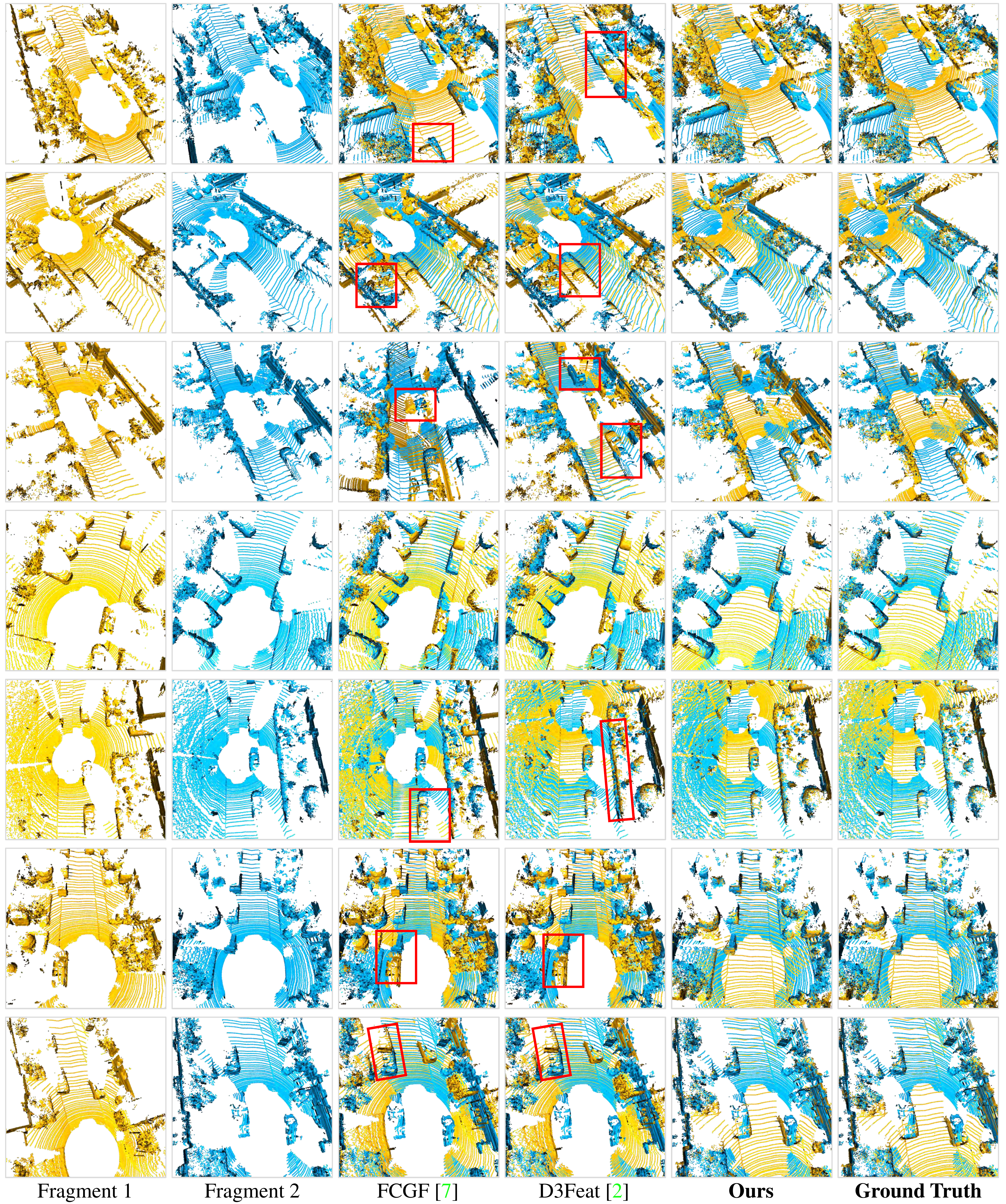}
	\end{center}
	\caption{\qysupp{Additional qualitative results achieved by FCGF \cite{choy2019fully}, D3Feat \cite{bai2020d3feat}, and our \textbf{\nickname{}} on the KITTI dataset. Note that, all methods are only trained on the indoor 3DMatch~\cite{Zeng2017} dataset.  Red boxes show the failure cases.}}
	\label{fig:supp_qualitative_KITTI}
\end{figure*}

\smallskip\noindent\textbf{Additional qualitative results on the KITTI dataset.} \qysupp{Then, we show the extra qualitative results achieved by FCGF \cite{choy2019fully}, D3Feat \cite{bai2020d3feat}, and our \nickname{} on the KITTI dataset in Fig. \ref{fig:supp_qualitative_KITTI}. We can clearly see that the point cloud in the KITTI dataset is significantly different from the point cloud in 3DMatch, since the KITTI dataset is mainly composed of \textit{large-scale}, \textit{sparse}, and \textit{partial} LiDAR scans. As shown in Figure, FCGF and D3Feat tend to misalign the input fragments when the scene contains lots of geometrically-similar objects (\textit{e.g.,} cars). However, our method can still achieve satisfactory registration results when only trained on the indoor 3DMatch dataset. This further validates the superior generalization ability of our method.}

\begin{figure*}[thb]
	\begin{center}
		\includegraphics[width=1.0\linewidth]{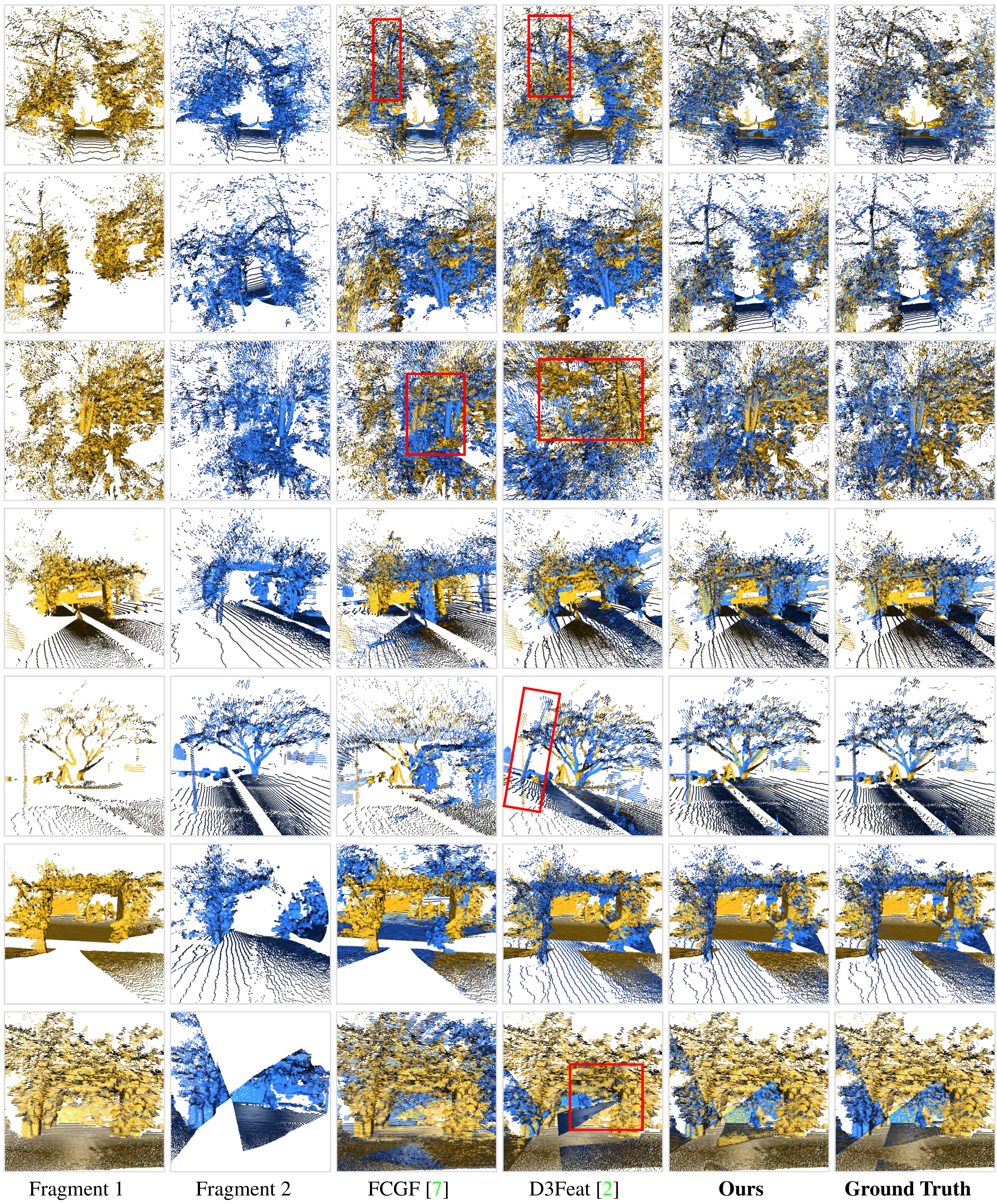}
	\end{center}
	\caption{\qysupp{Additional qualitative results achieved by FCGF \cite{choy2019fully}, D3Feat \cite{bai2020d3feat}, and our \textbf{\nickname{}} on the ETH dataset. Note that, all methods are only trained on the indoor 3DMatch \cite{Zeng2017} dataset.  Red boxes/circles show the failure cases.}}
	\label{fig:supp_qualitative_ETH}
\end{figure*}

\smallskip\noindent\textbf{Additional qualitative results on the ETH dataset.} \qysupp{We finally show the extra qualitative results achieved by FCGF \cite{choy2019fully}, D3Feat \cite{bai2020d3feat}, and our \nickname{} on the ETH dataset in Fig. \ref{fig:supp_qualitative_ETH}. Compared with the 3DMatch and KITTI data sets, the ETH dataset is collected by static terrestrial lasers in outdoor scenes, and is mainly composed of bushes and vegetation. As shown in Figure, it is highly challenging for FCGF and D3Feat to successfully align the input scans together, since this dataset suffers from issues such as noise, clutter, and occlusions. Nevertheless, the proposed \nickname{} can still achieve excellent performance on this dataset.}

% \begin{figure*}[thb]
% 	\begin{center}
% 		\includegraphics[width=1.0\linewidth]{figs/Fig_KITTI_supp2.pdf}
% 	\end{center}
% 	\caption{\qysupp{Additional qualitative results achieved by FCGF \cite{choy2019fully}, D3Feat \cite{bai2020d3feat}, and our \textbf{\nickname{}} on the KITTI dataset. Note that, all methods are only trained on the indoor 3DMatch~\cite{Zeng2017} dataset.  Red boxes/circles show the failure cases.}}
% 	\label{fig:supp_qualitative_KITTI2}
% \end{figure*}

% \begin{figure*}[thb]
% 	\begin{center}
% 		\includegraphics[width=1.0\linewidth]{figs/Fig_ETH_supp2.pdf}
% 	\end{center}
% 	\caption{\qysupp{Additional qualitative results achieved by FCGF \cite{choy2019fully}, D3Feat \cite{bai2020d3feat}, and our \textbf{\nickname{}} on the ETH dataset. Note that, all methods are only trained on the indoor 3DMatch \cite{Zeng2017} dataset.  Red boxes/circles show the failure cases, green boxes show the successful cases.}}
% 	\label{fig:supp_qualitative_ETH2}
% \end{figure*}

 \end{document}